%% file: main.tex
\documentclass[10pt,twocolumn,letterpaper]{article}

\usepackage{config/cvpr}

\input{config/definitions}
\input{config/preamble}

\input{config/authors}

\begin{document}

\title{\ourTitle}

\input{config/authors}
\input{paper/figs/teaser}

\input{paper/00_abstract}
\input{paper/01_intro}

\input{paper/02_related}

\input{paper/03_method}

\input{paper/04_experiments}

\input{paper/05_conclusion}

\input{paper/06_acknowledgments}

{
    \small
    \bibliographystyle{config/ieeenat_fullname}
    \bibliography{main.bbl}
}

\renewcommand{\thefigure}{S.\arabic{figure}}
\renewcommand{\thetable}{S.\arabic{table}}

\clearpage

\maketitlesupplementary

\input{supmat/01_video}
\input{supmat/02_grasp_transfer}
\input{supmat/021_runtime}
\input{supmat/022_physics_simulation}
\input{supmat/02_3_large_objects}
\input{supmat/04_qualitative_results}
\input{supmat/03_contact_heatmap}

\end{document}

%% file: config/definitions.tex
\usepackage{times}
\usepackage{epsfig}
\usepackage{graphicx}
\usepackage{amsmath}
\usepackage{amssymb}
\usepackage{caption}

\newcommand{\ourTitle}{\modelname: Generating Interaction Poses Using Spatial Cues and Latent Consistency}

\newcommand{\modelnameCOLOR}{black}

\newcommand{\modelname}{\textcolor{\modelnameCOLOR}{\mbox{GRIP}}\xspace}
\newcommand{\modelnameLong}{\textcolor{\modelnameCOLOR}{\mbox{Generating Realistic Interaction Poses}}}

\newcommand{\supmatCOLOR}{black}

\newcommand{\supmat}{\textcolor{\supmatCOLOR}{\textbf{Sup.~Mat.}}\xspace}

\usepackage{graphicx}
\usepackage{amsmath}
\usepackage{amssymb}
\usepackage{booktabs}
\usepackage{bm}
\usepackage{stmaryrd}
\usepackage{makecell}

\usepackage{balance}

\usepackage[pagebackref=true,breaklinks=true,colorlinks,bookmarks=false, pagebackref]{hyperref}

\usepackage{cuted}
\usepackage{enumitem}

\usepackage{multirow}
\usepackage{booktabs}
\usepackage{colortbl}
\usepackage{mathtools}

\newcommand{\changeCR}[1]{{\color{black} #1}}
\newcommand{\TODO}[1]{\textcolor{black}{#1}}
\newcommand{\rebut}[1]{\textcolor{black}{#1}}
\newcommand{\new}[1]{\textcolor{black}{#1}}
\newcommand{\REFINE}[1]{\textcolor{black}{#1}}

\newcommand{\colorRef}[1]{\textcolor{red}{#1}} %
\usepackage[capitalize]{cleveref}
\crefname{figure}{\colorRef{Fig.}}{\colorRef{Figs.}}
\Crefname{figure}{\colorRef{Figure}}{\colorRef{Figures}}
\crefname{section}{\colorRef{Sec.}}{\colorRef{Secs.}}
\Crefname{section}{\colorRef{Section}}{\colorRef{Sections}}
\Crefname{table}{\colorRef{Table}}{\colorRef{Tables}}
\crefname{table}{\colorRef{Tab.}}{\colorRef{Tabs.}}
\Crefname{equation}{\colorRef{Equation}}{\colorRef{Equations}}
\crefname{equation}{\colorRef{Eq.}}{\colorRef{Eqs.}}

\newcommand{\shape}{\bm{\beta}}

\newcommand{\pose}{\bm{\theta}}

\renewcommand{\etal}{et al.\xspace}
\renewcommand{\ie}{i.e.\xspace}

\newcommand{\grip}{\mbox{GRIP}\xspace}
\newcommand{\anet}{\mbox{CNet}\xspace}
\newcommand{\bnet}{\mbox{RNet}\xspace}

\newcommand{\toch}{\mbox{TOCH}\xspace}

\newcommand{\smplx}{\mbox{SMPL-X}\xspace}
\newcommand{\smplX}{\smplx}
\newcommand{\mano}{\mbox{MANO}\xspace}

\newcommand{\grab}{GRAB\xspace}

\newcommand{\grabnet}{\mbox{GrabNet}\xspace}

\newcommand{\mocap}{\mbox{MoCap}\xspace}

\newcommand{\threeD}{\xspace{3D}\xspace}

\newcommand{\groundtruth}{{ground-truth}\xspace}

\DeclareSymbolFont{matha}{OML}{txmi}{m}{it}%
\DeclareMathSymbol{\varv}{\mathord}{matha}{118}

\newcommand{\qheading}[1]{\noindent\textbf{#1:}}
\newcommand{\zheading}[1]{\textbf{#1:}}

\newcommand{\subject}[0]{}

\newcommand{\reals}[0]{\mathbb{R}}
\newcommand{\normabs}[1]{\lVert#1\rVert_1}
\newcommand{\normmse}[1]{\lVert#1\rVert_2}

\newcommand{\loss}[0]{\mathcal{L}}

\newcommand{\encoder}[0]{\mathcal{E}}
\newcommand{\decoder}[0]{\mathcal{D}}

\newcommand{\transl}[0]{\bm{\gamma}}

\newcommand{\vel}[1]{\dot{#1}}

\newcommand{\smplxparams}[0]{\bm{\Theta}^{\subject}}
\newcommand{\rhparams}[0]{\bm{\theta^r}}
\newcommand{\rhparamshat}[0]{\bm{\hat{\theta}^r}}
\newcommand{\raparamsi}[0]{\bm{{\theta}^{ra}}}
\newcommand{\raparamsp}[0]{\bm{{{\theta}}^{ra}_p}}

\newcommand{\lhparams}[0]{\bm{\theta^l}}
\newcommand{\lhparamshat}[0]{\bm{\hat{\theta}^l}}
\newcommand{\laparamsi}[0]{\bm{{\theta}^{la}}}
\newcommand{\laparamsp}[0]{\bm{{{\theta}}^{la}_p}}

\newcommand{\ambientsensor}[0]{\bm{h^A}}

\newcommand{\proximitysensor}[0]{\bm{{h}^P}}
\newcommand{\proximitysensorhat}[0]{\bm{\hat{h}^P}}

\newcommand{\handdist}[0]{\bm{d}}
\newcommand{\handdistrate}[0]{\vel{\bm{d}}}

\newcommand{\shapesubject}[0]{\shape^{\subject}}
\newcommand{\vertssubject}[0]{\verts^{\subject}}
\newcommand{\vertssubjectGT}[0]{\hat{\verts}^{\subject}}

\usepackage{soul}

\newcommand{\verts}[0]{\bm{v}}

\usepackage{lipsum}

\usepackage[normalem]{ulem} %

\usepackage{acronym}
\acrodef{amt}[AMT]{Amazon Mechanical Turk}    %

\usepackage{hhline}

%% file: config/preamble.tex
\usepackage[dvipsnames]{xcolor}

%% file: config/authors.tex
\author{
Omid Taheri\textsuperscript{1}\quad 
Yi Zhou\textsuperscript{3}\quad 
Dimitrios Tzionas\textsuperscript{2}\quad 
Yang Zhou\textsuperscript{3} \\
Duygu Ceylan\textsuperscript{3}\quad 
Soren Pirk\textsuperscript{3}\quad 
Michael J. Black\textsuperscript{1} \\ \\
\textsuperscript{1}Max Planck Institute for Intelligent Systems, 
\textsuperscript{2}University of Amsterdam,
\textsuperscript{3}Adobe Research\\
{\tt\small \textsuperscript{1}\{otaheri, tzionas, black\}@tue.mpg.de}
{\tt\small \textsuperscript{2}\{d.tzionas\}@uva.nl}\\
{\tt\small \textsuperscript{3}\{yzhou, yzhou2, dceylan, soren\}@adobe.com}
}

%% file: paper/figs/teaser.tex
\newcommand{\teaserCaption}{
        \rebut{\modelname generates realistic hand-object interaction poses (pink), given the easy-to-acquire body and object motion without fingers (blue) -- notice that the input hand pose is constant. \modelname animates the hands to be consistent with the body and object, producing realistic poses in various scenarios like pre-/post-grasp hand opening, and single or bi-manual grasps. It also works with various object shapes and sizes, and on different datasets like GRAB \cite{GRAB:2020} (left) and InterCap \cite{huang2022intercap} (right).}
        }

\newcommand\figcaption{\def\@captype{figure}\vspace{-10pt}\caption}

\twocolumn[
{
    \renewcommand\twocolumn[1][]{#1}
    \maketitle
    \centering
    \vspace{-0.8 em}
    \begin{minipage}{1.00\textwidth}
    \centering
	\includegraphics[trim=000mm 003mm 000mm 005mm, clip=true,width=1.0\linewidth]{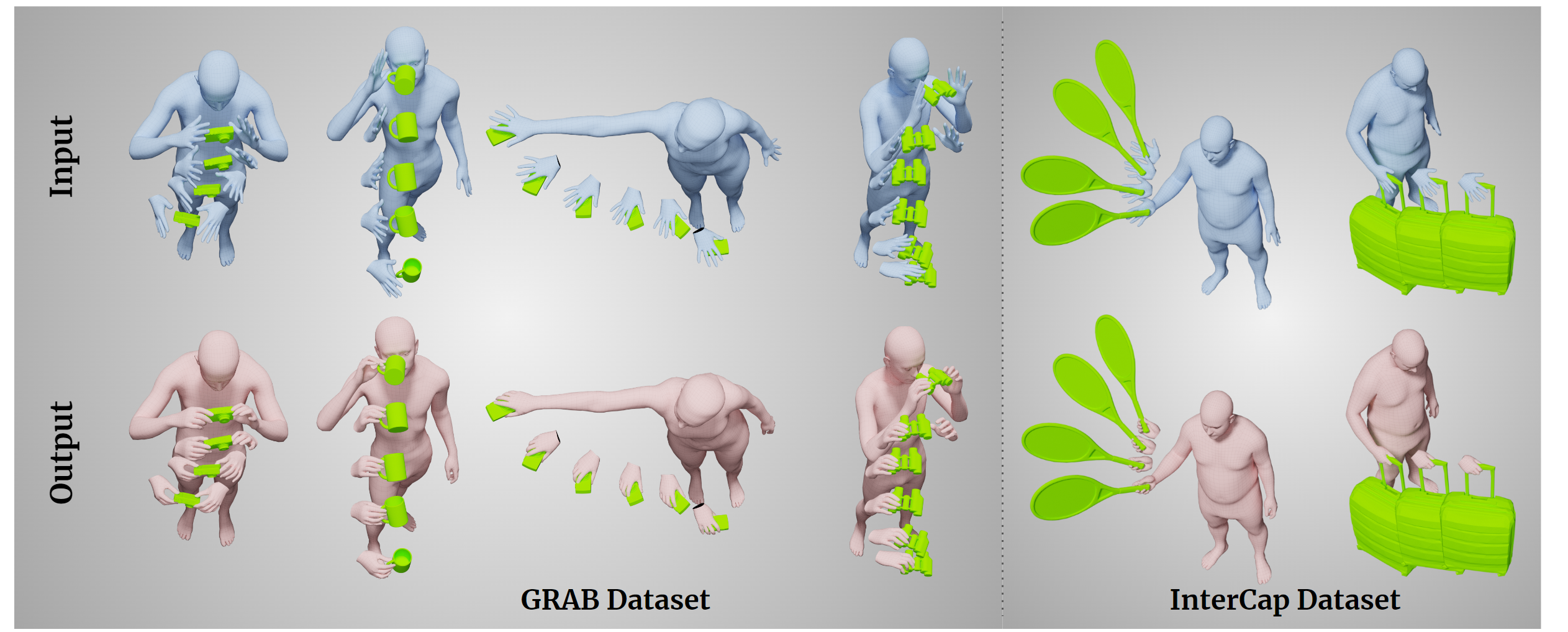}
    \end{minipage}
    \vspace{-1.0 em}
    \captionsetup{type=figure}
    \captionof{figure}{\teaserCaption}\label{fig:teaser}
    \vspace*{+01.50em}
}
]

%% file: paper/00_abstract.tex
\begin{abstract}
\vspace{-1.5 em}
Hands are dexterous and highly versatile manipulators that are central to how humans interact with objects and their environment.
Consequently, modeling realistic hand-object interactions, including the subtle motion of individual fingers, is critical for applications in computer graphics, computer vision, and mixed reality.
\rebut{Prior work on capturing and modeling humans interacting with objects in 3D focuses on the body and object motion, often ignoring hand pose.}
In contrast, we introduce \modelname, a learning-based method that takes, as input, the 3D motion of the body and the object, 
and synthesizes realistic motion for both hands before, during, and after object interaction.
\new{As a preliminary step before synthesizing the hand motion, we first use a network, ANet, to denoise the arm motion.}
Then, we leverage the spatio-temporal relationship between the body and the object to extract two types of novel temporal interaction cues, and use them in a two-stage inference pipeline to generate the hand motion. In the first stage, we introduce a new approach to enforce motion temporal consistency in the latent space (LTC), and generate consistent interaction motions.
In the second stage, \modelname generates refined hand poses to avoid hand-object penetrations.
\rebut{Given sequences of \new{noisy} body and object motion, \modelname ``upgrades" them
to include hand-object interaction.}
Quantitative experiments and perceptual studies demonstrate that \modelname outperforms baseline methods and generalizes to unseen objects and motions \rebut{from different motion-capture datasets.}
Our models and code \changeCR{will be} available for research purposes.

\end{abstract}

%% file: paper/01_intro.tex
\section{Introduction}	\label{sec:intro}
\input{paper/figs/grip_architecture}

Digital humans that move and interact naturally with 3D worlds have many applications in data creation, games, XR, and telepresence. 
In particular, physically plausible hand-object interaction is critical for realism. 
Unfortunately, automatically generating hand motions consistent with the world is challenging and no fully general solutions exist.

The problem is challenging since different object shapes require different types of interaction and hand grasps, such as a power grasp %
of an apple, a delicate three-finger pinching of a cup handle, and  %
bi-manual grasp of binoculars. Performing these actions is effortless for humans; however, even small errors, such as hand-object penetrations or subtly misplaced arms or fingers, can significantly affect the perceived realism of generated grasps for virtual avatars. 

Here we consider generating realistic grasps where a \threeD animation of the body and object is given, either from motion capture (MoCap), reconstructed from videos, or from an animator.
Motion capture data rarely contains hands as they are difficult to capture, requiring small markers that are often occluded and require high-resolution cameras. In some cases, despite being tracked, hands and arms are very noisy \cite{huang2022intercap}.
Objects, in contrast, are easier to track.
Figure \ref{fig:teaser} (top) illustrates this scenario with body and object motion, from GRAB \cite{GRAB:2020} and InterCap \cite{huang2022intercap}, but only rigid hands.
The goal is to transform this data into a more natural animation by synthesizing the appropriate hand-object interaction, as illustrated in Fig.~\ref{fig:teaser} (bottom).
With this approach we can ``upgrade'' existing datasets to support research on human-object interaction.

To this end, we introduce \emph{\modelname}, which stands for \emph{\modelnameLong}, a learned model that generates realistic hand motions for interactions with a variety of previously-unseen objects. Previous work in this direction focuses only on static grasping~\cite{GRAB:2020, grady2021contactOpt}, requires an initial hand pose that is then improved~\cite{zhou2022toch}, or only considers single-hand grasps~\cite{zhang2021manipnet, zhou2022toch}. Going beyond these approaches, our method  directly infers dynamic hand motion, \REFINE{both} in a single-hand or bimanual scenario, conditioned only on the object and body motion.

Our contributions are two-fold.
First,  we propose a set of virtual ``hand sensors" to extract rich spatio-temporal interaction cues between the body and the object.
Specifically, we introduce an \textit{Ambient Sensor} that senses the object shape and motion within the hands' \REFINE{broader} reaching region, as well as a \textit{Proximity Sensor} that captures fine-grained geometric features and a more nuanced distance field between the hand and object surface \REFINE{within the hands' closer} region. While virtual sensors have been used in prior work, our novel contribution is the innovative use of a distance-based representation combined with interaction-aware attention \cite{goal2022}.  This unique combination significantly improves results and generalizes to unseen objects and motions. 

Second, we propose \new{an \emph{arm denoising} network and} a novel two-stage \emph{hand inference} pipeline to leverage these features and generate realistic interaction motions. \new{Since arm motions from tracking or reconstruction can be noisy, we first use an arm denoising network, ANet to refine arm motion}. 
For the hand inference, our goal is to achieve near real-time performance, therefore, to avoid iterative optimization, like previous methods, we design two networks.
First, the \textit{Consistency Network (\anet)} takes features from both \textit{Hand Sensors} and generates smooth and consistent hand interaction motions.
Achieving this is challenging, as motions need to be realistic, temporally consistent, and natural. Naively applying  temporal smoothness terms to the final output hand motion, cf.~\cite{goal2022, starke2019neuralStateMachine}, will break the contact consistency and lead to high-frequency changes in contact areas. To overcome this, we propose a novel \textit{Latent Temporal Consistency (LTC)} solution. Specifically, we jointly learn global and residual latent codes to represent two successive frames and apply temporal consistency in the latent space, as shown in \cref{fig:architecture_Cnet}. Then, to mitigate any inconsistency between the two global latent codes, the key insight is to decode them using a ``shared'' network to generate consistent hand poses. \new{We use LTC in both ANet and CNet to ensure consistency in the motions. }

The generated hand poses from \anet bring fingers very close to the object surface, allowing the \textit{Proximity Sensor} to capture a more nuanced distance field. Therefore, in the second stage, we recompute the \textit{Proximity Sensor} features and use a refinement network, \bnet, to add subtle refinements and resolve penetrations in the interaction frames. 

\modelname is trained to generate both left- and right-hand motion simultaneously, enabling realistic modeling of single-hand %
and bi-manual interactions.
In contrast to other methods, which only focus on contact frames \cite{GRAB:2020, grady2021contactOpt}, our model is able to generate dynamic hand motions \emph{before}, \emph{during}, and \emph{after} the interaction with objects.
Additionally, unlike \cite{zhou2022toch}, which requires expensive optimization in the pose refinement step, our framework consists only of feed-forward neural networks.
By predicting realistic hand and finger motions, \modelname can be used to increase the realism of an avatar's interaction in AR/VR applications, refine noisy hand-object interaction motions (\cref{fig:intercap_results}\textcolor{red}{-left}), enrich existing interaction datasets that do not contain realistic finger motions (\cref{fig:intercap_results}\textcolor{red}{-right}), or capture new datasets with dexterous interactions but without explicitly tracking fingers. 

We evaluate \modelname quantitatively and qualitatively on a withheld test set from  the \grab dataset, with $5$ unseen objects and motions. The results show that our method generates accurate hand motions involving object grasping and manipulation. \rebut{We also show that \modelname generalizes to other MoCap datasets and larger objects, not present in \grab, by generating hand grasps for unseen objects from the MoGaze \cite{MoGaze2021} and InterCap \cite{huang2022intercap} datasets (see \cref{fig:intercap_results})}.
The quantitative evaluation shows that \modelname outperforms baselines, while our ablation studies explore the efficacy of our \textit{latent temporal consistency}, \textit{Hand Sensors}, and other design choices. Finally, we perform a perceptual study to evaluate  
the quality of the generated hand interaction motions. The results indicate that hand-object interaction sequences generated by \modelname achieve a level of realism similar to \grab's \groundtruth motions.

%% file: paper/figs/grip_architecture.tex
\begin{figure*}
    \vspace{-0.5 em}
    \centerline{
    \includegraphics[width=1.0\textwidth]{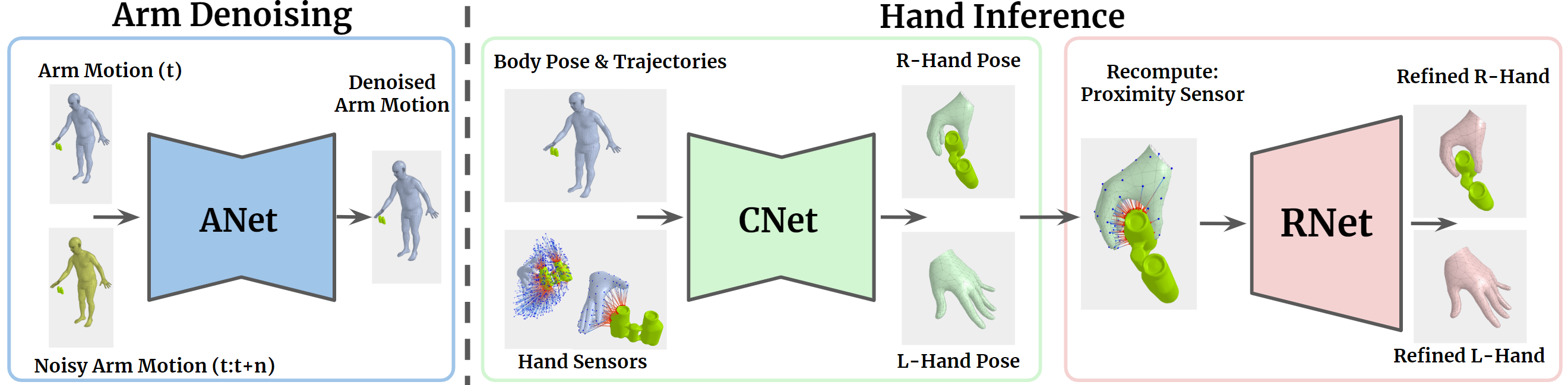}}
    \vspace{-1.5mm}
    \caption{
                Overview of \modelname.
                We first denoise the arm motion using the ANet network. We then predict hand interaction motion in two stages: (CNet)  Given the hand-object spatial features, extracted using our \textit{Hand Sensors}, body pose and trajectories in two consecutive frames, \anet  predicts both left- and right-hand poses.
                (RNet) Based on the predicted hand poses, we recompute the \textit{Proximity Sensor} feature and refine the hand poses with \bnet to enhance interaction accuracy and reduce possible penetrations.
    }
    \label{fig:grip_setup}
    \vspace{-0.5 em}
\end{figure*}

%% file: paper/02_related.tex
\section{Related Work}	\label{sec:related}
Despite the many advances in the field of motion synthesis for human avatars, generating accurate hand motion is still a challenging and unsolved problem. 
While many approaches focus on improving static grasps~\cite{GRAB:2020, grady2021contactOpt} with manually designed heuristics~\cite{Hasson2020PhotometricConsistency, hasson_2019_obman},
more recent techniques consider dynamic grasp generation~\cite{zhou2022toch, zhang2021manipnet}.
Such methods are still limited, and we review the most relevant ones below.

\zheading{Static Grasp Generation}
Generating static grasps has been widely studied in robotics, computer graphics, and computer vision. 
Common approaches in graphics and robotics use
physics-based control to generate novel hand grasps for a given 3D object. This includes using reference poses to optimize generated grasps \cite{pollard2005physically}, using hand pose and force closure  \cite{sahar2011garsp3d, kry2006interaction}, or pruning grasp candidates through physics-based analysis \cite{Bohg2014DataDrivenGrasp, li2007datadrivenGrasp, mordatch2012contactInvariant, Beatriz2010OpenGRASP}. Some recent methods take a data-driven approach and generate hand grasps by training on large hand-object interaction datasets \cite{GRAB:2020, Corona_2020_CVPR, contactDB_2019, Brahmbhatt_2020_ECCV, jiang2021graspTTA, Karunratanakul2021SkeletonDriven, karunratanakul2020graspField, Zhu2021DexterousGrasping}. Most of these approaches either estimate the grasping-hand pose directly~\cite{contactDB_2019, Brahmbhatt_2020_ECCV, jiang2021graspTTA}, based on model parameters \cite{SMPL:2015, smplifyPP} or by employing an implicit representation \cite{karunratanakul2020graspField, zhou2022toch}. Other approaches further refine the initially generated grasps by using a neural network \cite{GRAB:2020} or by leveraging predicted contact maps \cite{jiang2021graspTTA, grady2021contactOpt}.

\zheading{Dynamic Grasp Generation} 
Generating hand-object grasping motions is more challenging than static grasp generation. 
Most previous methods approach this by generating contact constraints and by resolving them through optimization-based methods \cite{liu2009dextrous, yeL2012synthesis, mordatch2012contactInvariant, zhao2013robustRealTime, liu2012synthesisHandonly}. Despite being physically plausible, the generated hand motions lack realism and are prone to interaction artifacts. More recently, reinforcement learning (RL) has been used for hand-only and full-body scenarios \cite{openAI2020rubic, rajeswaran2018learningComplex, Bergamin2019DReCon, Park2019UnorganizedMotion, peng2018deepmimic, Peng2017DeepLoco}. 
Christen et al.~\cite{christen2022dgrasp} employ physics simulation along with RL for dynamic grasp synthesis; however, their method requires reference hand-grasps and dynamic features of the object. A key challenge of these methods is generalization to new object geometries and hand configurations. Zhang et al.~\cite{zhang2021manipnet} use a distance-based spatial representation between hands and objects and train a network to generate right-handed object manipulation motions. 
To avoid interaction artifacts, \cite{zhou2022toch} propose an object-centric spatio-temporal representation and refine it with a neural network. The refined representation is then used in an optimization step to recover the hand-interaction motion. 
Unlike our approach, most of these methods treat each hand separately, making generated hand-collaboration and bi-manual grasps unrealistic.

\zheading{Object and Scene Interaction}
Some early work uses foot and hand contact annotations from \mocap datasets with optimization-based methods to extend or retarget human motions to scenes \cite{gleicher1998retargetting,lee2002interactive,lee2006motionPatches,kapadia2016precision}. 
Alternatively, deep reinforcement learning can be used to generate body-scene~\cite{chao2019learning2sit,peng2018deepmimic,peng2016terrain} or hand-object~\cite{christen2022dgrasp,GHernando2020physicsDext,chen2021systemHOI} interactions. 
Other methods use %
descriptors for dynamic interactions \cite{pirk2017understanding,Pirk2017SpatialTemporalDescriptors}, encode the joint motions of humans \wrt scene points \cite{alAsqhar2013relationship}, or use %
Laplacian deformation between body and object vertices to define a representation for modeling interactions~\cite{ho2010spatial}. As geometry-based approaches are not robust to real-world noise, some methods take a data-driven approach to predict action and motion sequences~\cite{Wang2019Interactions} or to %
generate key frames of motions and then complete them with data-driven or optimization-based techniques~\cite{hassan2021samp, goal2022,wang2021synthesizingLongTerm}.

\zheading{Hand-Object Interaction Tracking}
\REFINE{For graphics applications,} 
hand motions have traditionally been animated by artists~\cite{zhang2021manipnet}. 
While MoCap can be used to capture
hand motion datasets~\cite{contactDB_2019, Brahmbhatt_2020_ECCV, FirstPersonAction_CVPR2018, Shreyas2020HOnnotate, Tzionas:IJCV:2016}, such captures are technically challenging, limiting the amount of such data in the world.
For the MoGaze~\cite{MoGaze2021}, KIT~\cite{kitMocap2015}, and BEHAVE \cite{bhatnagar22behave}  datasets, human motions are tracked during interaction with %
objects, but the fingers and palm, are not explicitly captured. Taheri \etal \cite{GRAB:2020} capture accurate hand-object interactions with a high-accuracy \mocap system, but this approach 
does not 
scale. 
Zhang \etal \cite{zhang2021manipnet} propose a method for real-time hand motion synthesis, given the wrist and object motion. However, this %
does not %
generalize to new object shapes and full-body motions.  
InterCap \cite{huang2022intercap} captures full-body and hand interactions with objects, but hand poses are noisy.

\zheading{Summary}
Previous methods suffer from one or more of generalization ability, 
computation time, an initial hand pose requirement, or model only single-hand interactions. 
Our data-driven method, \modelname, addresses these limitations and  efficently generates realistic motions for both hands interacting with novel objects.

%% file: paper/03_method.tex
\section{Method}			
\label{sec:method}
Our goal is to add realistic hand poses to a body, based on the relative motion of the body and object during an interaction. 
To correctly estimate the hand interaction motion, we need to model how and when the object grasp happens. These cues can be found in the object's geometry and the correlated body-object motion trajectories. For example, if the distance between a wrist and the object is decreasing, the hand is approaching the object, but if it becomes constant and the object starts moving, we can infer it is grasped. %

To represent such information, we design two virtual ``hand sensors"; (1) the \textit{Ambient Sensor} obtains the object's geometric features and its spatial relation to the hands
and (2) the \textit{Proximity Sensor}  obtains a fine-grained distance field from different hand regions to the object surface. 

\new{However, if the arm motion is noisy, these computed features will also be inaccurate. Therefore, as a preliminary step, we use an arm denoising network, ANet, as shown in \cref{fig:grip_setup}, which takes the noisy arm motion and refines it while enforcing the temporal motion consistency.}

Then, we propose a two-stage hand prediction framework to generate hand motion, as illustrated in \cref{fig:grip_setup}. In the first stage, since we do not have an initial hand pose, we use a mean hand to compute the features of the hand sensors to predict both hand poses. To consider temporal information, we feed our model with the body poses and the hand sensors' features of the current and \TODO{next frame}, in addition to the hand-to-object distance and velocity in \rebut{the next $n$ frames (typically $10$, but this can be varied).}
In the second stage, based on the predicted hand poses, we recompute the \textit{Proximity Sensor} feature and refine the predictions to enhance interaction accuracy and reduce hand-object penetrations.
Details about each hand sensor and the neural networks are provided below.

\input{paper/figs/hand_sensor}

\subsection{Body and Hand Representations}		\label{sec:method_HumanModel}
To model the body and hand motion, we use the \smplX~\cite{smplifyPP} model. It can represent fine-detailed motion and accurate physical interactions, which are critical for object-interaction motions. Based on the body shape, $\shape$, and pose, $\pose$, parameters, \smplX reconstructs the body surface using linear blend skinning with a learned rigged skeleton, $\mathcal{J} \in \mathbb{R}^{55 \times 3}$. 
The full set of  \smplX parameters, $\smplxparams = \{\pose \in \reals^{55\times6}, \transl \in \reals^{3} \}$ includes both hands.
Here, we predict only the parameters of the hands:  the right-hand pose, $\rhparams \in \reals^{15\times6}$ and the left-hand pose, $\lhparams \in \reals^{15\times6}$ \cite{Zhou_2019_CVPR}. 
In addition, to efficiently represent the hand surface, we follow \cite{goal2022} and sample $99$ vertices on each hand; these are denoted $v^l$ and $v^r$, for the left and right hand, respectively.

\subsection{Ambient Sensor}
To sense the location and shape of the object,
we uniformly sample a set of $1024$ points in a hemisphere that is rigidly attached to each hand and centered at the base middle-finger joint, as shown in \cref{fig:hand_sensor}\textcolor{red}{-A}. For each motion frame, we compute the distance, $d$, from each of these points to the closest vertex on the object surface. 
This allows us to capture \TODO{detailed} information about the object shape and the relative distance between the hands and the object. The former informs the hand pose to adapt to certain shapes, while the latter helps predict the state of the hand motion, such as the pre-grasp and pre-release opening, and to keep consistent contact during the interaction.

Unlike commonly used voxel grids \cite{starke2019neuralStateMachine, zhang2021manipnet}, which provide a binary and discrete spatial representation, our novel \textit{Ambient Sensor} provides a continuous representation as it uses a distance-based representation. Furthermore, we pass the distances, $d$, through the interaction-aware attention transformation (\cref{eq:interaction_aware}) proposed by \cite{goal2022}, with $w=5$, to emphasize points closer to the object surface
\begin{equation}
     I_w(d) = \exp{\left(-w \times d\right)}, \quad w>0.
\label{eq:interaction_aware}
\end{equation}

\new{
The ablation studies in \cref{tab:grip_ablation_studies} and comparison with voxel-based ambient sensors show these unique combination captures rich spatial hand-object relations, improves results, and generalizes to unseen objects and motions. 
}

\subsection{Proximity Sensor}
Although the \textit{Ambient Sensors} capture important interaction information, they do not encode the distance of specific hand regions to the surface of the object; this is essential to know the contact areas. Therefore, we use the sampled hand vertices $\verts$ and compute their closest distance to the object surface. Since we do not have the hand pose in the beginning, we initialize the hand with the mean pose from SMPL-X~\cite{smplifyPP} and compute the proximity features in the first stage of prediction, as shown in \cref{fig:hand_sensor}\textcolor{red}{-B}. In the second stage, we recompute \emph{Proximity Sensors}' values using the hand poses generated from the first stage, as shown in \cref{fig:hand_sensor}\textcolor{red}{-C}.

In contrast to the \textit{Ambient Sensor}, the \textit{Proximity Sensor} provides fine-grained geometric details. This more nuanced information about interaction is essential to generate hand poses with fewer penetrations and better contacts, particularly when the hands are very close to the object's surface. Thus, \REFINE{for the \textit{Proximity Sensor}}, we apply the transformation in \cref{eq:interaction_aware} with a higher weight ($w=50$) \wrt the 
\emph{\REFINE{Ambient} 
Sensor}, to put emphasis on the vertices closer to the object.

\subsection{Consistency Network (\anet)}
\label{sec:Cnet}
\anet is a novel encoder-decoder neural network that takes the body motion and hand sensor features of two consecutive frames at time $t$ and $t+1$ to predict the hand poses of both frames. The two frames will be used in our proposed \textit{Latent Temporal Consistency (LTC)} algorithm to enforce temporal and contact consistency for the final prediction.
\anet additionally takes the average hand-to-object distance $\handdist$ in the future $n$ frames, from $t$ to $t+n$, where $n=10$ by default, as input to better disambiguate the grasp and release moments. The detailed architecture of \anet is illustrated in \cref{fig:architecture_Cnet}.
The inputs to the network are:
\begin{equation}
X = 
\left[\shapesubject, {\pose}_{t:t+1}, \ambientsensor_{t:t+1}, \proximitysensor_{t:t+1},
\Bar{\handdist}_{t:t+n}, \Bar{\handdistrate}_{t:t+n}
\right]
\label{eq:anet_enc_inputs}
\end{equation}
where 
$t:t+i$ denotes $i$ motion frames in the future including the current frame, 
${\pose}_{t:t+1}$ are the \smplX joint angles without considering the global root joint,
$\ambientsensor_{t:t+1}$ and $\proximitysensor_{t:t+1}$ are the hand \emph{\REFINE{Ambient} Sensor} and \emph{Proximity Sensor} values for both left and right hands,
and ${\Bar\handdist}_{t:t+n}$ and $\Bar{\handdistrate}_{t:t+n}$ are the average of  hand-to-object distance and its rate of change for sampled hand vertices in the $n$ future frames.

\input{paper/figs/cnet_architecture}

\zheading{Latent Temporal Consistency (LTC)}
In addition to physically plausible hand-object contact, an important factor 
in the realism of 
interaction motions is consistent dynamics and contact areas between consecutive frames. 
To enforce these, 
we  smooth the motion in the latent space 
of hand motions rather than in the output space, as we noticed the latter adds high-frequency noise to the contact areas throughout the motion. 
As shown in \cref{fig:architecture_Cnet}, the encoder, $\encoder^{\text{C}}$, maps the input $X$ to two latent codes, $z_t, z_{t+1}^t \in \reals^{256}$, where ${z_t}$ denotes the global latent code for a hand pose in the current frame and $z^t_{t+1}$ is the relative latent code for the next frame with respect to the current frame. We compute the global latent code for the next frame by adding the two latent codes as ${z_{t+1} = z_t + z_{t+1}^t}$; see \cref{fig:architecture_Cnet}. We then pass each global latent code individually to a shared decoder, $\decoder^{\text{C}}$, to get the outputs $\hat{Y}$. The shared decoder helps %
regulate inconsistency between the two global latent codes, as it 
is 
represented and penalized in the final hand poses.
The output of \anet is:
\begin{equation}
{\hat{Y}} = 
\left[{\rhparamshat}_{t:t+1}, {\lhparamshat}_{t:t+1}, {\proximitysensorhat}_{t:t+1},
\right]
\label{eq:anet_outputs}
\end{equation}
where ${\rhparamshat}_{t:t+1}$, ${\lhparamshat}_{t:t+1} \in \reals^{15\times6}$ are right-/left-hand poses in the current and next frame, and ${\proximitysensorhat}_{t:t+1}$ are the inferred \emph{Proximity Sensor} values; 
\REFINE{the latter ones have been shown to increase realism and lower errors \cite{goal2022}.} 

Generating hand poses in the current and next frame allows for defining consistency and smoothness losses between them. Evaluations in \cref{tab:grip_ablation_studies} show that the motions generated with our LTC algorithm achieve a lower error and better consistency compared to baselines with no enforced consistency or with consistency in the output space.

We use fully-connected dense residual blocks with skip connections for both the encoder and decoder, and train \anet end-to-end.
The training loss is defined as 
\begin{align}
\loss = 
\lambda_{\verts} \loss_{\verts}            +
\lambda_{\proximitysensor} {\loss_{\proximitysensor}} +
\lambda_{\theta} {\loss_{\theta}} 
\label{eq:anet_train_loss},
\end{align}
where
$\loss_{\verts}              = \normabs{\vertssubject - \vertssubjectGT}$ is a loss on the hand vertices $\verts$,~
$\loss_{\theta}             = \normmse{\lhparamshat  - \lhparams} + \normmse{\rhparamshat  - \rhparams}$ is on the joint rotations of both hands and
$\loss_{\proximitysensor}    = \normabs{\REFINE{\proximitysensorhat}  - \proximitysensor}$ is on the hand-to-object distances, both directly estimated from the network and derived from the estimated hand poses.

\new{\subsection{Arm Denoising Network (ANet)}
For the hand sensors in \anet to capture rich information between the hand and the object, the motion of these two should be  very accurate and without noise. Therefore, as shown in \cref{fig:grip_setup} (left), we train ANet to first refine the arm motion before passing to \anet. It takes as input both arms' pose in the current frame, $\laparamsi$ and $\raparamsi$, and the noisy poses of the future frame,  $\laparamsp$ and $\raparamsp$, and gives the denoised arm poses. We use a similar architecture to \anet, and enforce the consistency between the denoised poses in the latent space of the network using LTC. For more details about ANet please see \supmat
}

\subsection{Refinement Network (\bnet)}
\label{sec:Rnet}
The motions generated by \anet are 
\REFINE{in the right ballpark} 
but can be refined further to improve realism and remove possible penetrations. To this end, we train a refinement network, \bnet. We use the generated hand poses from \anet to recompute \textit{Proximity Sensor} features, $\proximitysensor_{\theta}$, similar to \anet inputs (see \cref{fig:hand_sensor}\textcolor{red}{-C}). Then \bnet takes $\proximitysensor_{\theta}$ and the hand poses, $\lhparamshat$ and $\rhparamshat$, and outputs the \TODO{refined hand poses}. To keep the motion dynamics, generated from \anet, we train \bnet to refine hand poses only in the interaction frames and not to change the pose when hands are far away from the object surface. In addition to the \anet output, we train \bnet on perturbed training data to %
simulate noisy inputs. Training losses are similar to those used for \anet in \cref{eq:anet_train_loss}. \bnet consists of 3 fully-connected residual layers with skip connections in between, for an architectural overview, more details, and the data processing pipeline please see \supmat.

%% file: paper/figs/hand_sensor.tex
\begin{figure}
    \centering
    \includegraphics[trim=000mm 000mm 000mm 000mm, clip=true, width=1.00 \linewidth]{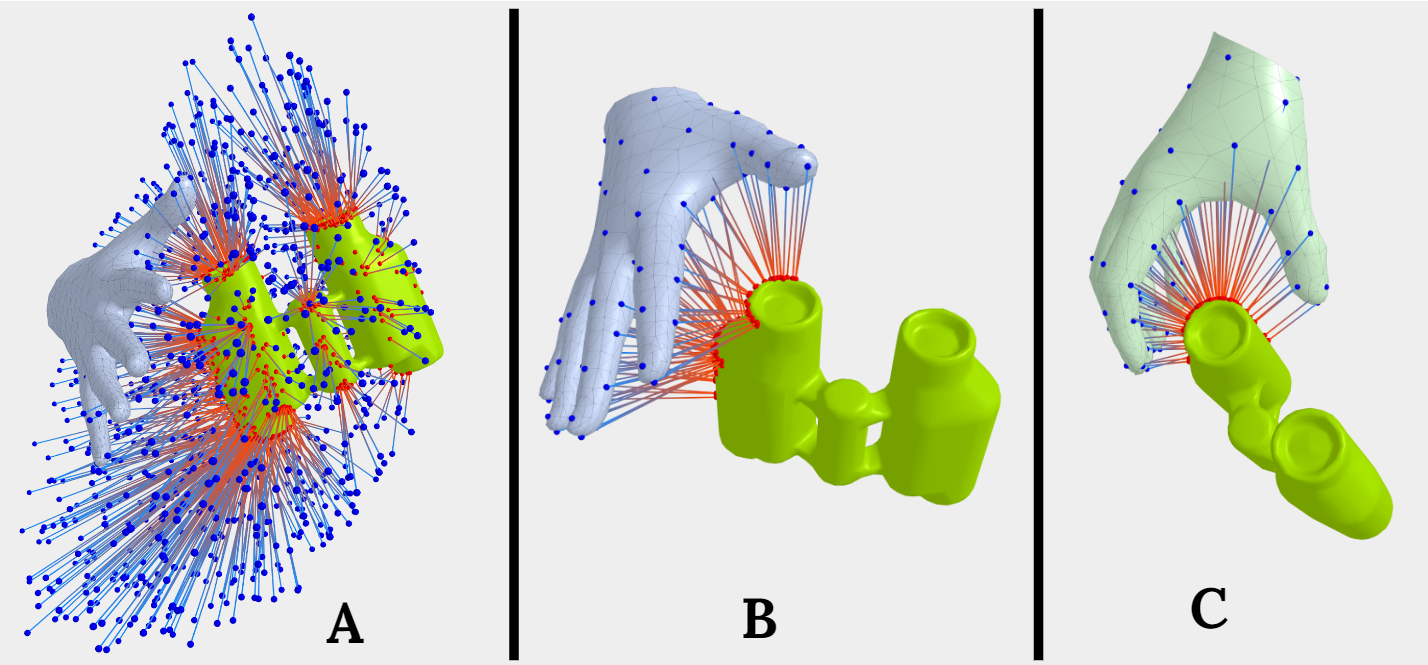}
    \caption{
            Visualization of our Hand Sensors (only right-hand for simplicity).
            \textbf{(A)} \emph{Ambient Sensor} points (blue) and their computed distances to the closest object points (red). This captures the object geometry and distance to the hands.
            \textbf{(B)} \emph{Proximity Sensor} feature computation for  \anet's inputs with mean-hand pose initialization. 
            \textbf{(C)} Recomputing the Proximity Sensor values for \bnet, using the hand poses generated by \anet. Note that the corresponding points on the object change for each finger compared to (B).
    }
    \label{fig:hand_sensor}
    \vspace{-0.5 em}
\end{figure}

%% file: paper/figs/cnet_architecture.tex
\begin{figure}[bt]
    \centering
    \includegraphics[trim=006mm 000mm 000mm 000mm, clip=true, width=1.00 \linewidth]{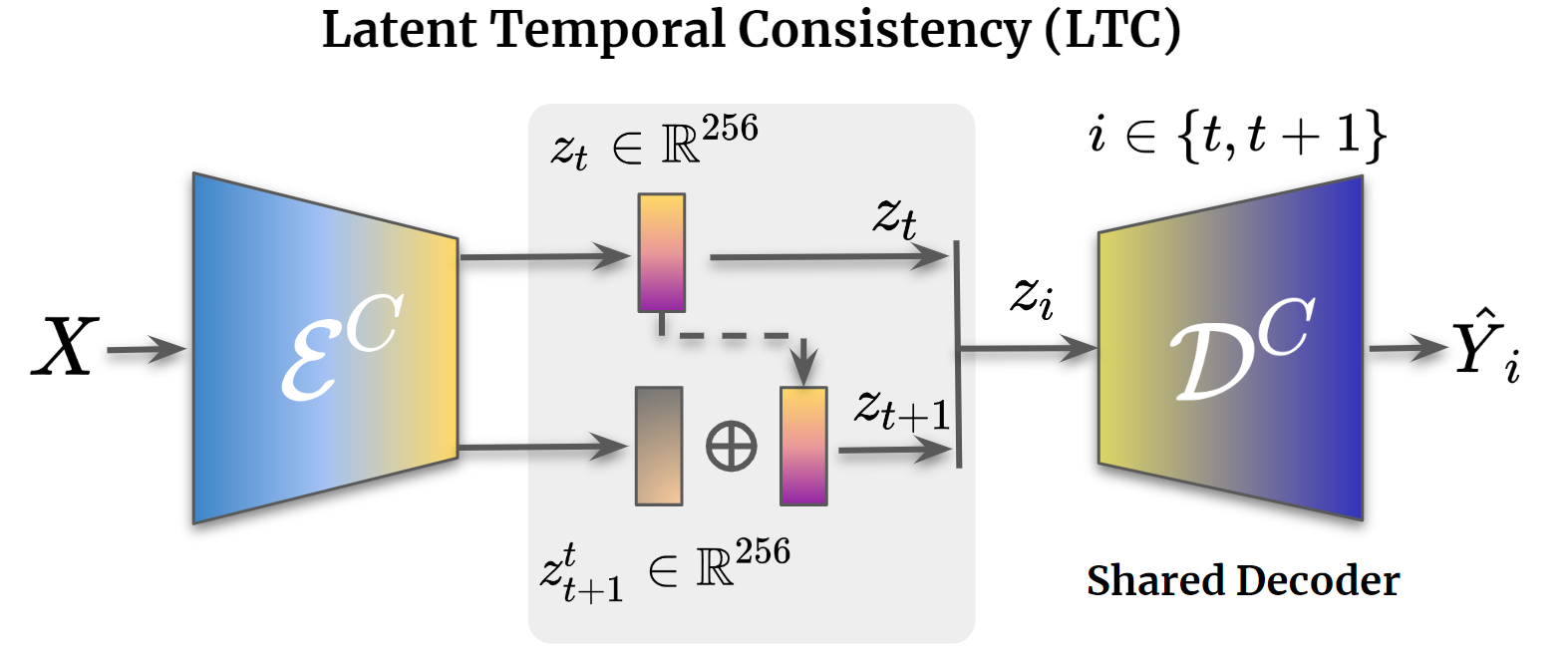}
    \caption{
            \anet Architecture. We propose the LTC algorithm that enforces consistency between two successive frames in the latent space (see \cref{sec:Cnet} for more details).
    }
    \label{fig:architecture_Cnet}
\end{figure}

%% file: paper/04_experiments.tex
\section{Experiments}  \label{sec:experiments}
\input{paper/figs/rnet_cnet_compare}

\subsection{Evaluation Metrics}
We use the standard %
``Mean Per-Joint Position Error'' (MPJPE) and ``Mean Per-Vertex Position Error'' (MPVPE), which represent the Euclidean distance between the \groundtruth and estimated hand joints and vertices, respectively.

\qheading{Intersection Volume (IV)} This measures the intersection volume between the hand and the object to assess the realism, \REFINE{\ie, the physical plausibility,} of the generated grasps.

\qheading{Contact Consistency (CC)} This evaluates the consistency of contacts for the grasping frames of 
generated grasp motions, \REFINE{\ie,} the \TODO{finger sliding} on the object surface. 
We use \groundtruth motions to select grasp frames, and, for generated motions, compute the deviation distance from the contact areas on the object. 

\subsection{Qualitative Evaluation}	\label{sec:experiments_qualitative}

Results show that \anet generates reasonable and smooth hand grasps, but sometimes with artifacts like hand-object interpenetration. After applying the refinement network, \bnet, the results look more realistic and physically plausible. In \cref{fig:rnet_cnet} we show examples of generated grasps using \anet and  after applying the \bnet refinement.

\Cref{fig:teaser,fig:results_qualitative} show several representative hand motions generated with \modelname, including pre-/post-grasp hand opening, single-hand grasps, and bi-manual grasps for different unseen object shapes. Overall, the generated hand motions are reasonable, smooth, and consistent.
For more results, please see \supmat

\smallskip
\qheading{Performance on Other Datasets}
\rebut{
\modelname is trained on the GRAB dataset, which only has small hand-held objects. 
High-quality data of hand-object interaction with large objects is rare.
Despite training on small objects, our virtual hand sensors help generalize to larger objects, as they only sense the interaction areas \REFINE{locally} and not the whole object. 
To highlight \modelname's generalization capability, we show generated interaction poses for \emph{unseen} large objects from the InterCap \cite{huang2022intercap} and MoGaze \cite{MoGaze2021} datasets in \cref{fig:intercap_results} and \cref{fig:teaser}\textcolor{red}{-right}, and compare them with the original hand poses.
For more results, see \supmat}

\smallskip
\qheading{\REFINE{Cross-Object} Grasp Transfer}
\rebut{We show that \modelname can be used to transfer grasping motions from one object to another one, for the details and results please see \supmat} 

\input{paper/figs/intercap_results}

\input{paper/figs/results_qualitative}

\subsection{Ablation Study}  \label{sec:experiments_ablations}

\qheading{Latent Temporal Consistency (LTC)} 
To evaluate the importance of our proposed temporal consistency algorithm for interaction motions, we compare our network with two baselines, namely: 
(1)~a network without enforced consistency (\emph{w/o Consist.}) and 
(2)~a network with consistency applied directly on the generated hand poses (\emph{output Consist.}). 
As seen in \cref{tab:grip_ablation_studies}\textcolor{red}{-bottom} our LTC method that smooths the latent space representation not only reduces the CC error, but also results in  lower errors in  MPVPE and MPJPE.

\input{paper/tabs/GRIP_evaluation_qualitative}

\input{paper/tabs/GRIP_evaluation_ablations}

\smallskip
\qheading{Hand Sensors} 
To evaluate the effect of our \emph{Ambient Sensor} and \emph{Proximity Sensor}, we train different baselines of \modelname by removing these features, (\emph{w/o Ambient}) and (\emph{w/o Proximity}), and additionally compare them to Voxel-based representation. 
We compare MPVPE, MPJPE, and CC between the generated hand motions and the ground truth. 
Results in \cref{tab:grip_ablation_studies}\textcolor{red}{-top} show that our distance-based hand sensors provide rich interaction information to the network that leads to lower errors and consistent motions. 

\smallskip
\qheading{\bnet} In \cref{tab:grip_ablation_studies} we evaluate our refinement network, \bnet, by comparing the results of \modelname with \bnet \REFINE{(\emph{fullmodel})} and without it \REFINE{(\emph{w/o \bnet})}. The table verifies that the refinement step helps reduce the hand MPJPE and MPVPE errors and enhance motion consistency. 

\smallskip
\qheading{Number of Future Frames}
\rebut{In \cref{tab:evaluations_rebuttal}\textcolor{red}{-right} we compare different variants of \modelname to show the effect of using a different number of future motion frames on the accuracy
of the generated hand poses. 
The table verifies using more future frames (up to 10 frames) lets the network generate more accurate poses. 
This is a trade-off between a \TODO{real-time} performance (row~1) and a higher accuracy with some latency \mbox{(rows~2-4)}.} 
\REFINE{Empirically, we observe that performance saturates for more than 10 frames, in accordance with \cite{goal2022}.} For details on the inference runtime, please see \supmat
\input{paper/tabs/GRIP_evaluation_comparison}

\subsection{Perceptual Study (Comparison to ManipNet)}
\new{We evaluate the hand motions generated from \anet and \bnet with a perceptual study on \ac{amt} and compare them with ManipNet results and the GT motions.
For \REFINE{GRAB's} test-set motion sequences, we use \grip to generate the interacting hand poses. We then create videos of the generated motions from \anet, the refined motions from \bnet, and the corresponding ground truth. To compare with ManipNet, we extracted their moving meshes from their demo and rendered them in the same format as \grip results.}

\new{
The participants rate the realism of the hand motions based on $4$ criteria:
(1)     hand-object grasp, 
(2)     hand motion smoothness,
(3)     contact consistency, and
(4)     in-hand manipulations.
Each motion is evaluated by at least $10$ different participants. The ratings are on a 5-level Likert scale, where $1$ means unrealistic and $5$ means very realistic.
We use a catch trial similar to \cite{GRAB:2020, goal2022} to identify invalid \TODO{ratings} and remove \TODO{them}; \cref{tab:grip_results_amt} shows the evaluation results.} 

\new{
The study shows that the \modelname-generated hand motions are very realistic and close to the \groundtruth ones. In addition, the scores are slightly higher 
\REFINE{when motions are} 
refined by \bnet, especially for Contact Consistency (CC), which shows the effectiveness of our LTC algorithm. 
Furthermore, we see a lower rating for ManipNet results compared to our results. 
Additionally, in \cref{tab:grip_ablation_studies} we show the computed penetration errors for ManipNet, which is $13$\%~higher than ours. 
While the test data is different (simpler for ManipNet), these results confirm several limitations of ManipNet such as single-hand inference, poor generalization to new objects, and no full-body setting.
GRIP addresses these issues, making it easy to apply in real-world scenarios. For representative grasps and failures please see \supmat.
}

\subsection{Comparison to TOCH}
\new{
To evaluate the performance of ANet and \bnet, we compare them to TOCH \cite{zhou2022toch} on refining perturbed test-sets from \grab.
To do this, similar to \cite{zhou2022toch}, we perturb the motions by adding Gaussian noise, with different magnitudes, to the pose (GRAB-R) and translation (GRAB-T) of both hands.
To keep the original motion dynamics, generated from \anet, \bnet is trained to only refine hand-pose (\ie,~rotation perturbations), therefore we refine perturbed translation using ANet and perturbed rotations using \bnet. We provide the full-comparison results in \cref{tab:grip_comparison_toch}. 
Results show that the combination of ANet and \bnet performs better in refining noisy hand interactions.}

\input{paper/tabs/merged_evaluations}

\subsection{Baselines}
\new{
To evaluate \grip's performance, in \cref{tab:evaluations_rebuttal}\textcolor{red}{-left} we compare 
the penetration volume ($\text{cm}^3$) and contact ratio 
\cite{zhang2020generating} 
of two GrabNet variants and ManipNet with our models.  namely:
(1) ``\grabnet'' \cite{GRAB:2020}, which generates \mano grasps, 
(2) a trained ``\grabnet-\smplx'' variant, which generates full-body \smplx grasps, 
(3) ManipNet,
(4) \modelname (w/o \bnet), and 
(5) \modelname (w/ \bnet). 
Results show that our full model (row~5) performs better than baselines in generating realistic grasps. 
Please note that our model generates the \emph{motion} of \emph{both hands}
during object interaction with realistic transitions between no-grasp poses, pre-grasp openings, grasping, and releasing of objects, while ManipNet generates single-hand motions, and GrabNet variants only generate \emph{static} grasps of \emph{one} hand.
}

%% file: paper/figs/rnet_cnet_compare.tex
\begin{figure}[bt]
    \centering
    \includegraphics[trim=006mm 000mm 008mm 000mm, clip=true, width=1.00 \linewidth]{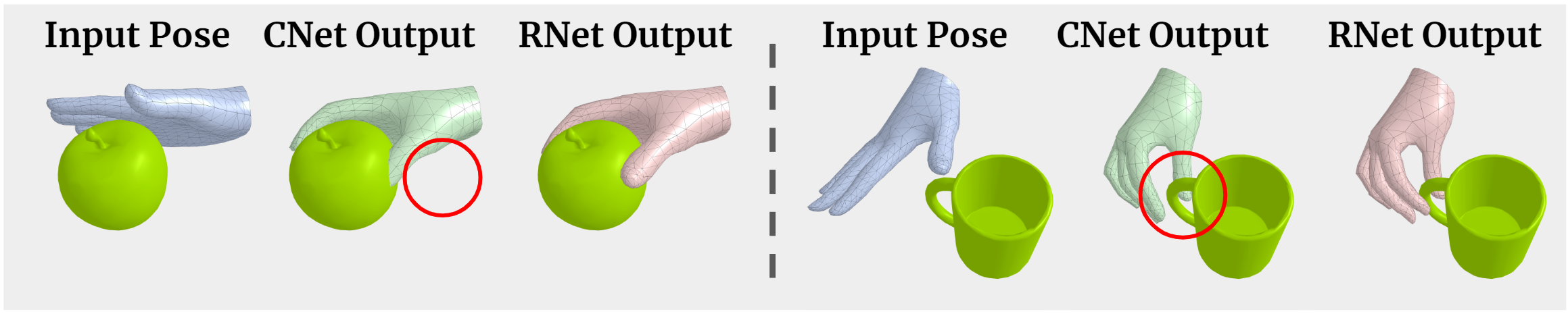}
    \caption{
            Comparing \anet and \bnet generated grasps.
            Results show that \bnet effectively refines the penetration and ``non-contact" artifacts (red circles) of the \anet results.
    }
    \label{fig:rnet_cnet}
\end{figure}

%% file: paper/figs/intercap_results.tex
\begin{figure}[t!]
    \centering
    \includegraphics[trim=009mm 000mm 005mm 000mm, clip=true, width=1.00 \linewidth]{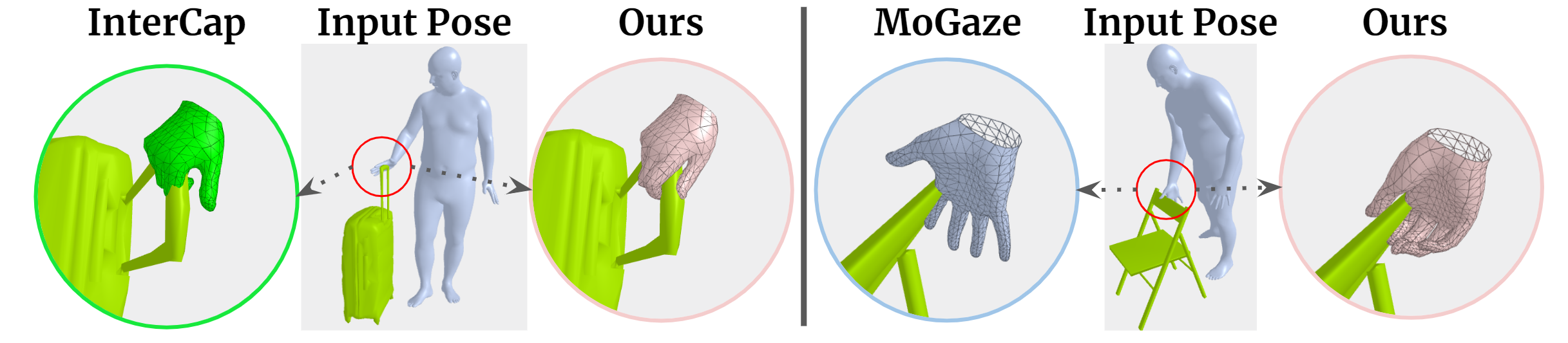}
    \caption{Our generated grasps (\REFINE{pink circles})
    for %
    large objects from InterCap 
    \cite{huang2022intercap} 
    and MoGaze \cite{MoGaze2021}
    , and
    comparison with the original grasps from these datasets.
    }
    \label{fig:intercap_results}
\end{figure}

%% file: paper/figs/results_qualitative.tex
\begin{figure*}[t]
	\centering			%
	\includegraphics[trim=000mm 090mm 000mm 000mm, clip=true,width=1.0\linewidth]{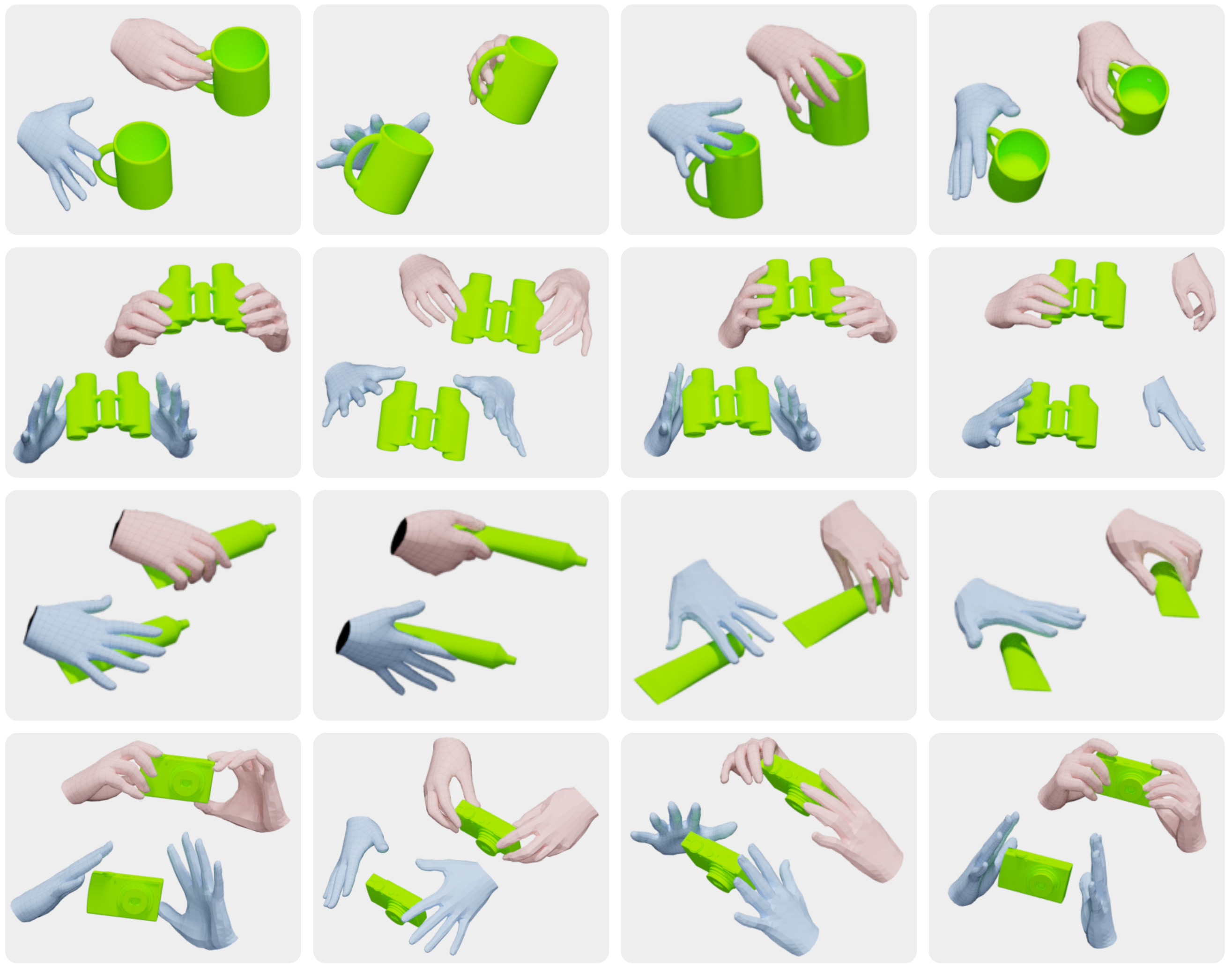}
    \caption{
                \modelname results. 
                We show various generated grasps, in single and bimanual scenarios, for different objects shapes.
                \REFINE{The input (flat, non-articulated) hands are shown with blue meshes, and \modelname's generated hands (articulated) with pink  meshes.}
    }
	\label{fig:results_qualitative}
 \vspace{-0.5em}
\end{figure*}

%% file: paper/tabs/GRIP_evaluation_qualitative.tex
\begin{table}[tb!]
\centering
\renewcommand{\arraystretch}{1.2}
\small
\resizebox{1.00\linewidth}{!}{
\begin{tabular}{l|cccc}
\Xhline{3\arrayrulewidth}
\multicolumn{1}{l}{\textbf{Metric}} & \textbf{ManipNet} & \textbf{\grip (w/o \bnet)}  & \textbf{\grip} & \textbf{Ground truth} \cite{GRAB:2020}
\\
\hline
Hand-Object Grasp   $\uparrow$ & $3.68 \pm 1.05 $ & $4.09 \pm 0.89 $  & {$4.11 \pm 0.85 $}     & $4.12 \pm 0.90 $  \\ 
Hand Motion Smoothness     $\uparrow$ & $3.8 \pm 0.93 $& $3.88 \pm 1.06 $  & {$3.91 \pm 1.04 $}     & {$3.98 \pm 1.03 $}  \\ 
Contact Consistency $\uparrow$ & $3.54 \pm 0.99 $ & $4.02 \pm 1.01 $  & {$4.09 \pm 0.95 $}     & {$4.13 \pm 0.95 $}  \\ 
In-Hand Manipulation      $\uparrow$ & $3.57 \pm 0.99 $ & $3.96 \pm 1.01 $  & {$3.97 \pm 0.99 $}     & $4.01 \pm 1.00$   \\ \hline
Average               $\uparrow$ & $3.65 \pm 1.00 $ & $3.99 \pm 1.00 $  & {$4.02 \pm 0.96 $}     & $4.06 \pm 0.97$   \\
\Xhline{3\arrayrulewidth}
\end{tabular}
} 
\vspace{-3mm}
\caption{%
    Perceptual evaluation of \modelname results, without and with \bnet, compared with the ManipNet \cite{zhang2021manipnet} results and ground truth \cite{GRAB:2020}. The participants rate the realism of the generated grasps from $1$ (unrealistic) to $5$ (very realistic).
    The table reports the mean $\pm$ std, computed for all valid study participants.
    Results show that \grip generated grasps are more realistic than ManipNet and that \bnet improves the grasps of \anet.
}
\vspace{-4mm}
\label{tab:grip_results_amt}
\end{table}

%% file: paper/tabs/GRIP_evaluation_ablations.tex
\begin{table}
\centering
\renewcommand{\arraystretch}{1.2}
\small
\resizebox{1.\linewidth}{!}{
\begin{tabular}{lcccccc}
\Xhline{3\arrayrulewidth} 
\multicolumn{1}{l|}{\multirow{2}{*}{\textbf{Method} $\downarrow$}} & \multicolumn{2}{c}{\textbf{MPVPE}   (mm) $\downarrow$} & \multicolumn{2}{c}{\textbf{MPJPE}   (mm) $\downarrow$} & \multicolumn{2}{c}{\textbf{CC} (mm) $\downarrow$} \\ \cline{2-7} 
\multicolumn{1}{l|}{}                                   & R-Hand        & \multicolumn{1}{c|}{L-Hand} & R-Hand        & \multicolumn{1}{c|}{L-Hand} & R-Hand                     & L-Hand                    \\ \Xhline{2\arrayrulewidth}
                                                      & \multicolumn{6}{c}{\textbf{Hand   Sensors Ablation}}                                                                                               \\ \hline
\multicolumn{1}{l|}{\modelname(w/o   Ambient)}          & 9.56          & 6.72                        & 7.08          & 4.99                        & 15.03                      & 9.48                      \\
\multicolumn{1}{l|}{\modelname(w/o   Proximity)}        & 9.62          & 6.82                        & 7.11          & 5.09                            & 15.64                      & 9.10                      \\ \Xhline{2\arrayrulewidth}
                                  & \multicolumn{6}{c}{\textbf{Latent  Temporal Consistency \REFINE{(LTC)} Evaluation}}                                                                               \\ \hline
\multicolumn{1}{l|}{\modelname(w/o   Consist.)}         & 8.17          & 6.18                        & 5.99          & \textbf{4.53}               & 13.01                      & 7.66                      \\
\multicolumn{1}{l|}{\modelname(output   Consist.)}      & 9.31          & 7.11                        & 6.81          & 5.31                        & 13.21                      & 8.18                      \\
\multicolumn{1}{l|}{\new{\modelname(Voxel-grid)}}          & 8.36          & 6.54                        & 6.60          & 4.75                        & 11.35                      & 6.87                      \\
\Xhline{3\arrayrulewidth}
\multicolumn{1}{l|}{\modelname (w/o \bnet)}            & 8.19          & 6.58                        & 6.10          & 4.95                        & 11.44                      & 7.03                      \\
\multicolumn{1}{l|}{\modelname   (fullmodel)}          & \textbf{7.88} & \textbf{6.17}               & \textbf{5.85} & 4.62                        & \textbf{10.56}             & \textbf{6.25}             \\ \Xhline{3\arrayrulewidth} 
\end{tabular}
} 
\vspace{-0.5 em}
\caption{%
    \textbf{(Top)} 
    We show the effect of our ``Hand Sensors'' by comparing variants of \modelname without our sensors' features; \modelname results in lower errors.
    \textbf{(Bottom)} 
    The effect of the LTC algorithm is explored by comparing \modelname against a network without LTC (\emph{w/o Consist.}) and one with consistency on the output poses (\emph{output Consist.}). 
    The \modelname-generated motions have lower errors.
}
\label{tab:grip_ablation_studies}
\vspace{-0.5 em}
\end{table}

%% file: paper/tabs/GRIP_evaluation_comparison.tex
\begin{table}
\centering
\renewcommand{\arraystretch}{1.2}
\small
\resizebox{1.00\linewidth}{!}{
\begin{tabular}{l|c|c|c|c|c}

 \Xhline{3\arrayrulewidth}

\multicolumn{1}{l|}{Metric $\downarrow$} &
  \multicolumn{1}{c|}{Model $\downarrow$} &
  \multicolumn{1}{c|}{\begin{tabular}[c]{@{}c@{}}GRAB-T\end{tabular}} &
  \multicolumn{1}{c|}{\begin{tabular}[c]{@{}c@{}}GRAB-T\end{tabular}} &
  \multicolumn{1}{c|}{\begin{tabular}[c]{@{}c@{}}GRAB-R\end{tabular}} &
  \multicolumn{1}{c}{\begin{tabular}[c]{@{}c@{}}GRAB-R\end{tabular}} \\ %
\multicolumn{1}{l|}{} &
  \multicolumn{1}{c|}{} &
  \multicolumn{1}{c|}{\begin{tabular}[c]{@{}c@{}}(0.01)\end{tabular}} &
  \multicolumn{1}{c|}{\begin{tabular}[c]{@{}c@{}}(0.02)\end{tabular}} &
  \multicolumn{1}{c|}{\begin{tabular}[c]{@{}c@{}}(0.3)\end{tabular}} &
  \multicolumn{1}{c}{\begin{tabular}[c]{@{}c@{}}(0.5)\end{tabular}} \\\Xhline{2\arrayrulewidth}
\multirow{2}{*}{MPVPE (mm)} & TOCH & 16.0 $\rightarrow$ \textbf{11.8} & 31.9 $\rightarrow$ \textbf{13.9} & \textbf{6.30} $\rightarrow$ 11.5 & \textbf{10.3} $\rightarrow$ 11.0 \\ \cline{2-2}
                       & GRIP & 17.4 $\rightarrow$ \textbf{10.3} & 34.2 $\rightarrow$ \textbf{13.1} & 6.21 $\rightarrow$ \textbf{4.62} & 10.5 $\rightarrow$ \textbf{6.72} \\ \hline
\multirow{2}{*}{MPJPE (mm)} & TOCH & 16.0 $\rightarrow$ \textbf{9.93} & 31.9 $\rightarrow$ \textbf{12.3} & \textbf{4.58} $\rightarrow$ 9.58 & \textbf{7.53} $\rightarrow$ 9.12 \\ \cline{2-2}
                       & GRIP & 16.9 $\rightarrow$ \textbf{9.70} & 33.8 $\rightarrow$ \textbf{12.8} & 4.26 $\rightarrow$ \textbf{3.21} & 7.64 $\rightarrow$ \textbf{4.18} \\ \Xhline{3\arrayrulewidth}
                       
\end{tabular}

}
\vspace{-0.5em}
\caption{%
    Comparison of \modelname (ANet and \bnet) performance with \toch \cite{zhou2022toch} on the perturbed test-sets from \grab. Following \toch, we perturb the hand pose (-R) and translation (-T) by adding Gaussian noise. 
    \REFINE{The numbers in parentheses (top) show the noise magnitude.}
    The table reports the metrics before and after using each method. 
}
\vspace{-0.5em}
\label{tab:grip_comparison_toch}
\end{table}

%% file: paper/tabs/merged_evaluations.tex
\newcolumntype{?}{!{\vrule width 1.5pt}}
\begin{table}[t!]
\vspace{-0.2 em}
\centering
\small
\renewcommand{\arraystretch}{1.1}
\footnotesize
\resizebox{1.0\columnwidth}{!}{

\begin{tabular}{l|c|c?c|c|c}
\cmidrule(r){1-3} \cmidrule(l){4-6}
\multicolumn{1}{c}{Grasp }     & Penetr.                    & Cont.                & \multicolumn{1}{c}{\multirow{1}{*}{\modelname}} & \multicolumn{2}{c}{MPVPE} \\ 
\multicolumn{1}{c}{ Synthesis} & ($\text{cm}^3$) $\downarrow$ & Ratio  $\uparrow$    & \multicolumn{1}{c}{\multirow{1}{*}{\# Future}}         & \multicolumn{2}{c}{(mm) $\downarrow$} \\ 
 \cline{1-3} \cline{5-6} 
1. \grabnet 
    \cite{GRAB:2020} 
    & 2.65 & \textbf{1.00}    & \multicolumn{1}{c}{Frames}  &   R-Hand & L-Hand  \\  
\cline{4-6}
2. \grabnet-\smplx            & 7.33 & 0.87   & 1.~~~ 0~~ & 9.21 & 8.18         \\
3. ManipNet & 2.68 & 0.98 & 2.~~~ 3~~ & 8.94 & 7.78 \\
3. \modelname (w/o-\bnet)    & 3.18 & 0.96          & 3.~~~ 5~~ & 8.34 & 7.29 \\
\cellcolor{blue!25}{4. \modelname (w/~~-\bnet) } &
  \textbf{2.38} &
  \textbf{1.00}          & \cellcolor{blue!25}{4.~~~ 10} &
  \textbf{7.88} &
  \textbf{6.17} \\ \cline{1-3}
~~~~~\grab (GT)                & 1.95                 & 1.00 &
   &
   &
     \\ 
\cmidrule(r){1-3} \cmidrule(l){4-6}
\end{tabular}
}
\vspace{-.5 em}
\caption{
    \rebut{
    \textbf{(Left)}
    Penetration and contact-ratio metrics for two
    \grabnet
    baselines and \modelname models.
    \textbf{(Right)}
    Evaluating the trade-off for the real-time performance and accuracy of \modelname by comparing different numbers of future frames. 
    }
}
\vspace{-0.6 em}
\label{tab:evaluations_rebuttal}
\end{table}

%% file: paper/05_conclusion.tex
\section{Conclusion}                        \label{sec:discussion}
We propose \modelname, a data-driven method that directly generates realistic interaction motions for both hands given the animated body and target object. Our method's novelties include (1) an arm denoising network and a two-stage hand prediction approach using two networks for coarse and fine grasping, (2) the combination of two novel distance-based hand sensors, and (3) a latent-space temporal consistency modeling. As a result, compared with previous methods, \modelname is able to refine noisy interaction motions and then predict hand poses from scratch, generalize to novel object shapes, adapt to bi-manual interactions, and generate realistic hand poses with temporal consistency. These benefits will allow \modelname to be used for capturing new datasets of human-object interaction {\em without} the difficulty of tracking the hands, to add hands to previous datasets~\cite{MoGaze2021, kitMocap2015}, and to synthesize hands for avatars in video games and AR/VR.

\zheading{Limitations and Future Work}
Although the inference time for \modelname is very fast, it relies on mean hand-to-object distance in the future $10$ frames to guide the prediction of grasps.  This causes a fixed 10-frame latency in \TODO{interactive} applications. It may be possible to learn to anticipate movement and reduce this delay.
Extending the method to human-scene interaction would be interesting.

%% file: paper/06_acknowledgments.tex
\smallskip

\qheading{\emph{Acknowledgements}}
This research has been started during Omid Taheri's internship at Adobe Research and is a collaboration with the Max Planck Institute for Intelligent Systems. It was partially supported by Adobe Research and the International Max Planck Research School for Intelligent Systems (IMPRS-IS),, and the German Federal Ministry of Education and Research (BMBF).
We thank Tsvetelina Alexiadis for the Mechanical Turk experiments. \\
\qheading{\emph{Disclosure}}
MJB has received research gift funds from Adobe, Intel, Nvidia, Meta/Facebook, and Amazon.  MJB has financial interests in Amazon, Datagen Technologies, and Meshcapade GmbH.  While MJB is a consultant for Meshcapade, his research in this project was performed solely at, and funded solely by, the Max Planck Society.

%% file: supmat/01_video.tex
In this supplemental material, we provide additional information about \modelname as mentioned in the main paper; this includes details of the method, more qualitative results, grasp analysis, and the details of the cross-grasp transfer application. 

\section{Data Preparation}				\label{sec:method}
The \grab dataset \cite{GRAB:2020} is used to train our \modelname model.  It is a \mocap dataset that accurately captures whole-body motions involving the manipulation of  \threeD objects. The body is parameterized with \smplX~\cite{smplifyPP}.
The motions are performed by $10$ participants on $51$ objects with different shapes and sizes. We withhold 5 objects for the test-set and use the rest for training and validation of the networks. 
\qheading{\anet data}
\anet generates hand interaction motion based on the body and object motion in a sequence. 
We use all the training and test sequences from \grab for training and testing \anet, respectively.
In addition to hand-object grasp frames, we consider other motion frames of each sequence to generalize to pre-grasp and post-grasp hand poses. 
In total, we use $1335$ motion sequences, performed on $51$ {3D} objects. To split the dataset, we use the motions performed on ``mug", ``apple", ``camera", ``binoculars", and ``toothpaste" as the test set,``fryingpan'', ``toothbrush'', ``elephant'', and ``hand'' as the validation set, and the rest as the training set. In total, we have $329K$, $52K$, and $24K$ motion frames for the training, testing, and validation set, respectively.

\qheading{\bnet data}
\bnet refines the motions generated from \anet, therefore, we use the output of \anet as the main data source for \bnet. In addition, to model more severe penetration and interaction artifacts, we prepare a synthetic dataset by perturbing the \groundtruth data in \grab. For this, we add Gaussian noise with a standard deviation of $0.3$ to the axis-angle rotation representation of the hand poses.

\qheading{ANet data}
ANet is trained to refine noisy arms motion. To prepare the training data, we add Gaussian noise to the shoulder and elbow joints of the \groundtruth motion data. The noise is added to the axis-angle rotation of the joints and has $0.01$ and $0.03$ standard deviations for the shoulder and elbow joints, respectively. 

\input{supmat/figs/anet_arch}
\section{Arm Denoising Network (ANet)}

For an architectural overview of ANet see \cref{fig:anet_ltc}. As input, A-Net takes the arm motion and hand sensor features of the current Ground Truth frame along with five noisy future frames. As output it gives the denoised arm poses for the five future frames, following \cite{goal2022}. To ensure motion consistency between the successive frames of the denoised motions, we use the LTC algorithm similar to \anet, as explained in the main manuscript (Sec. 3.4).  For this, the encoder, $E^A$, maps the input to five latent representations for each arm pose, as shown in \cref{fig:anet_ltc}. Then we apply the latent temporal consistency algorithm by adding the residual latent codes, $z^t_i$, to the global latent code, $z_t$. Finally, we use a shared decoder, $D^A$, to decode the denoised motions. Both encoder and decoder have $4$ fully-connected residual layers with skip connections in between.

\section{\bnet Network}
For the architecture overview of \bnet please see \cref{fig:rnet_arch}. \bnet takes, as input, hand poses and proximity sensor values of a motion frame and, as output, generates the refined hand poses for both left and right hand. The network consists of $4$ residual blocks with skip connections and an output linear layer.
\input{supmat/figs/rnet_arch}

%% file: supmat/figs/anet_arch.tex
\begin{figure*}
	\centering			%
	\includegraphics[trim=000mm 000mm 000mm 000mm, clip=true,width=1.0\linewidth]{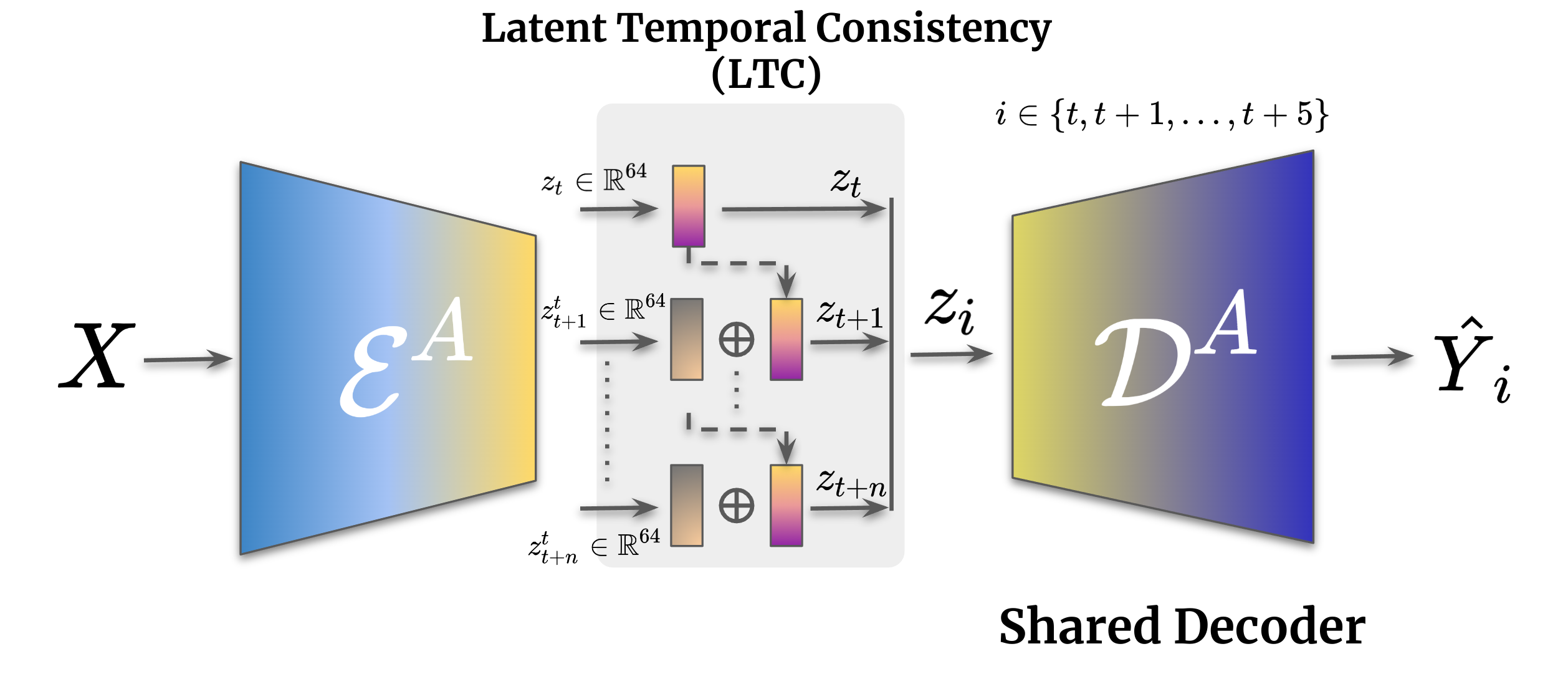}
    \caption{Architecture overview of ANet. Similar to CNet, we use the LTC algorithm to ensure motion consistency of the denoised arm motions. For this, the encoder maps the input to a global latent code in the current frame and relative latent codes in the future frames. Then a shared decoder is used to generate the denoised motions.
    }
	\label{fig:anet_ltc}
\end{figure*}

%% file: supmat/figs/rnet_arch.tex
\begin{figure*}
	\centering			%
	\includegraphics[trim=000mm 000mm 000mm 000mm, clip=true,width=1.0\linewidth]{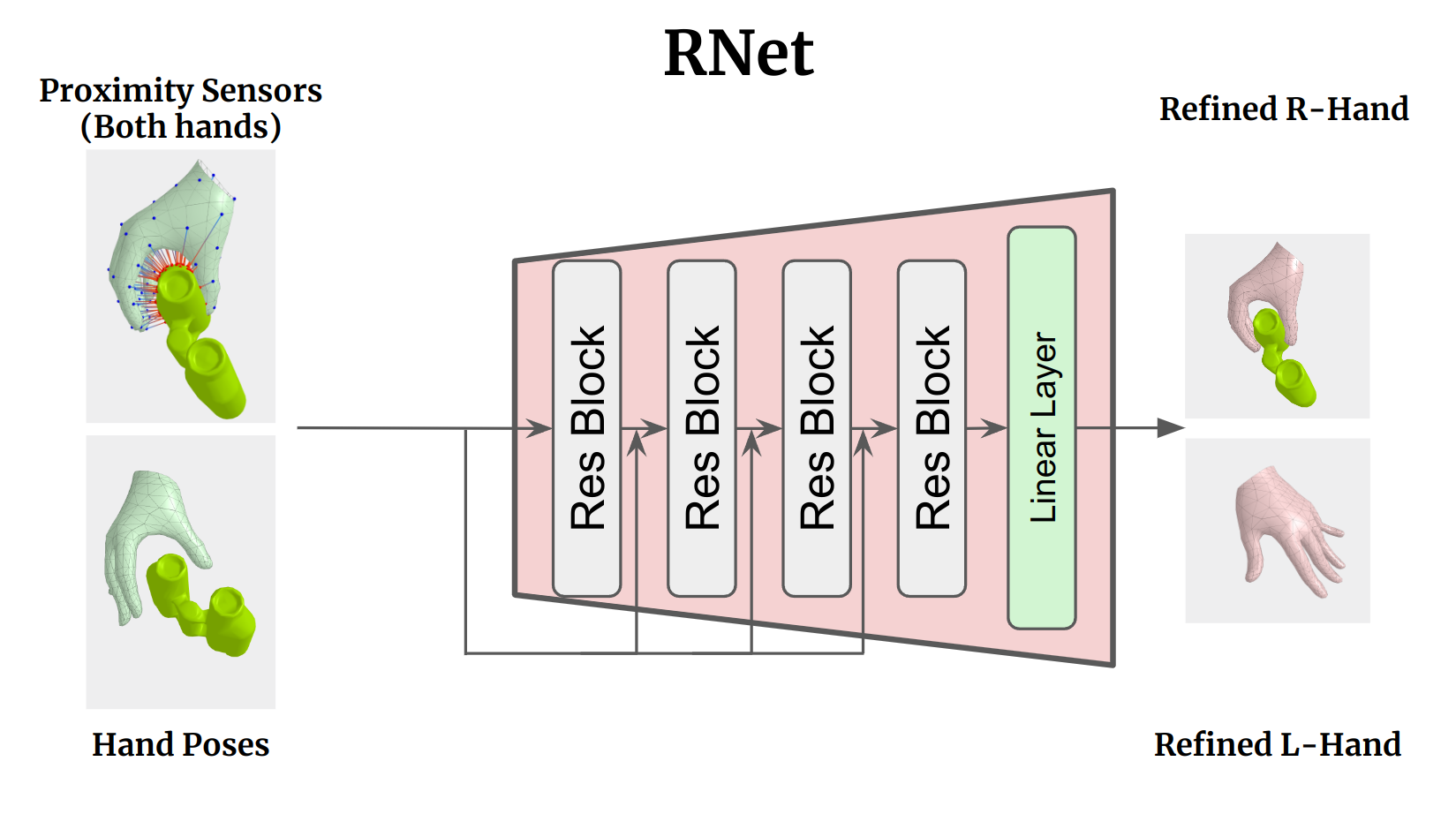}
    \caption{Architecture overview of \bnet. As input, it takes hand poses and proximity sensor values, and generates the refined hand poses. The network consists of $4$ residual blocks with skip connections and an output linear layer.
    }
	\label{fig:rnet_arch}
\end{figure*}

%% file: supmat/02_grasp_transfer.tex
\section{Grasp Transfer (Application)} \label{sec:grasp_transfer}
To test whether our method generalizes well to different object shapes and motions, we use \modelname to transfer the input interaction motion from a source object to a target object. Given a sequence of body and object motion without hand poses, we replace the source object with a target object that is roughly of the same size. We then compute the hand sensor features for the new object geometry and use \modelname to generate hand interaction poses for the new object. 

Qualitative results show that our method is able to generate realistic hand motions for the target object and generalizes well to the new object's shape and motion. In \cref{fig:grasp_transfer} we show two examples of the grasp transfer application. The top row shows that the hands adapt well to the target object geometry, ``elephant'', and the bottom row shows a change in the grasp type (e.g., thumb contact area) due to the smaller size of the target object, ``sphere''.  
This is useful for synthetic data generation because a single motion capture sequence can be repurposed to generate many different synthetic human-object interactions.
This is also useful for FX where actors are captured handling a ``dummy" object that is replaced by a 3D graphics object; this is a common scenario in film production.

\input{supmat/figs/grasp_transfer}

%% file: supmat/figs/grasp_transfer.tex
\begin{figure*}[hbt!]
	\centering			%
	\includegraphics[trim=000mm 000mm 000mm 000mm, clip=true,width=1.0\linewidth]{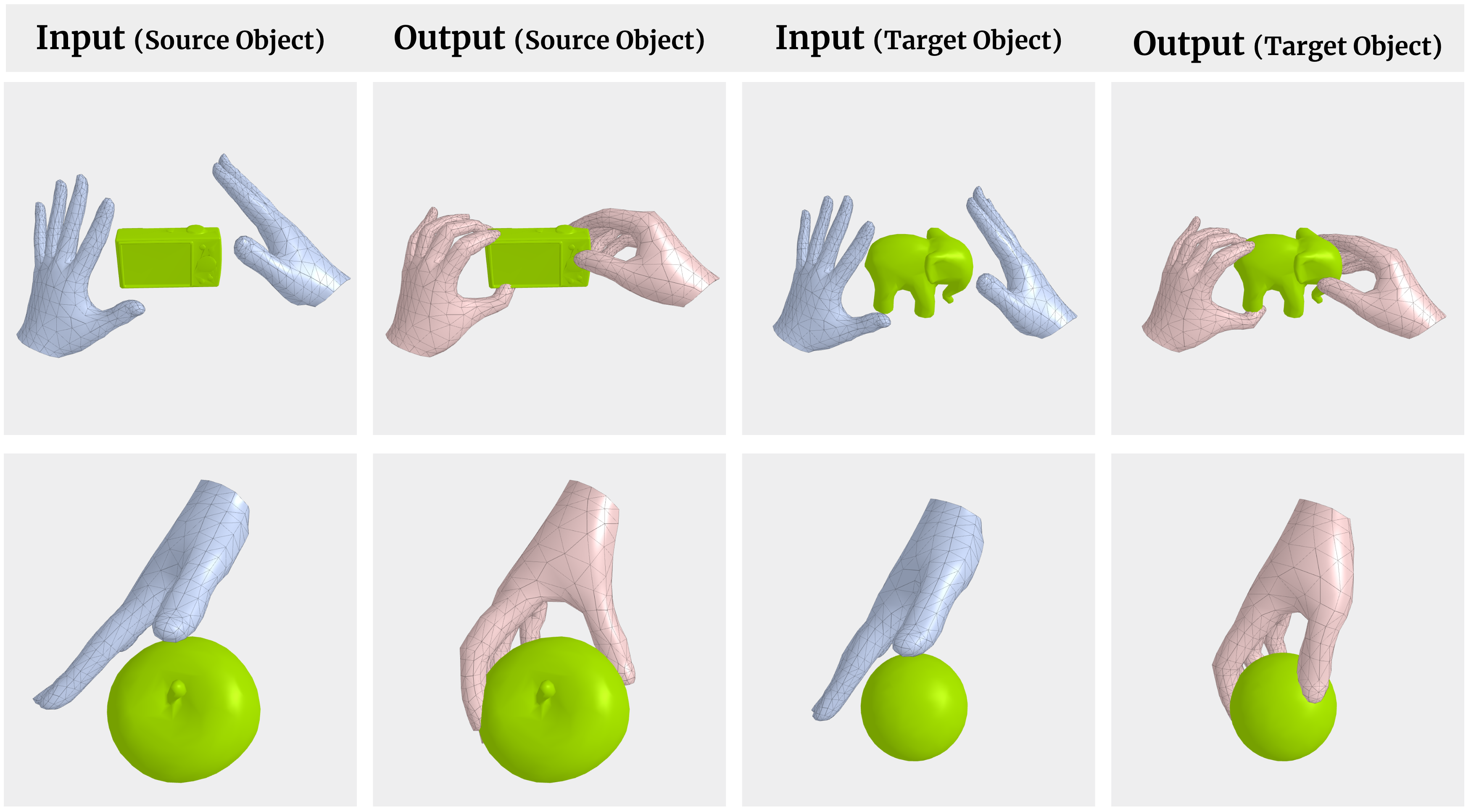}
    \caption{Grasp transfer from a source object to a target one.  Given a sequence of body and object motion without hand poses, we replace the source object with a target one and use \modelname to generate hand interaction poses for the new object. The top row shows grasp transfer from ``camera'' to an ``elephant'' geometry, and the bottom row shows grasp transfer from an ``apple'' to a small ``sphere''. Notice how the hands adapt to the new object shape (top row) and the change in the grasp type (bottom row).
    }
	\label{fig:grasp_transfer}
\end{figure*}

%% file: supmat/021_runtime.tex
\section{Runtime}

Due to its pure learning-based pipeline, \modelname is able to generate hand poses rapidly. 
We find that a full forward pass of our method (without ANet) on a single V100-16GB GPU, including the \anet inference, recomputing proximity sensor values, and \bnet forward pass, takes $0.022$ seconds, which is equivalent to $45$ fps. 
Therefore, \modelname can be used to synthesize hands for avatars in interactive applications like video games and mixed reality settings, which are mostly running at $30$ fps. Please notice that our network still relies on mean hand-to-object distance in the future $10$ frames, which causes a fixed 10-frame latency ($1/3$ of a second) in real-time applications. \rebut{This is the trade-off to have more accurate poses with latency instead of real-time performance with lower accuracy, as shown in \textcolor{red}{Tab. 4-right} in the main paper.}

%% file: supmat/022_physics_simulation.tex
\section{Physics Simulation}

Our main goal is to generate visually plausible hand-object interaction motions, however we also evaluate the physical plausibility of our results, which may be important for the real-world applications.
Following prior methods \cite{hasson_2019_obman, Hasson2020PhotometricConsistency, karunratanakul2020graspField}, 
we evaluate the generated grasps in a Bullet physics simulation.
We fix the body position and apply gravity to the object. A small object displacement (\textless 1 mm) after 5 physics simulation steps is counted as a ``stable'' grasp.
For all generated grasps, \anet and \bnet have $93\%$  and  $97\%$  stability, respectively.
This suggests that the synthesized hand poses are not just visually pleasing but also physically realistic.

%% file: supmat/02_3_large_objects.tex
\section{Performance on Large Objects} \label{sec:large_objs}
\input{supmat/figs/intercap_mogaze_results}
In \cref{fig:large_objects} we show more qualitative results of our method performance to generate hand grasps for large objects. 
Note that these objects have extended 3D structure compared with all the training objects in the GRAB dataset.
What is important to note here is that our hand sensors are not distracted by the extended objects due to their locality.
Thus \modelname is able to generate plausible grasps for such objects.

%% file: supmat/figs/intercap_mogaze_results.tex
\begin{figure*}[hbt!]
    \centering			%
    \includegraphics[trim=000mm 000mm 000mm 000mm, clip=true,width=1.0\linewidth]{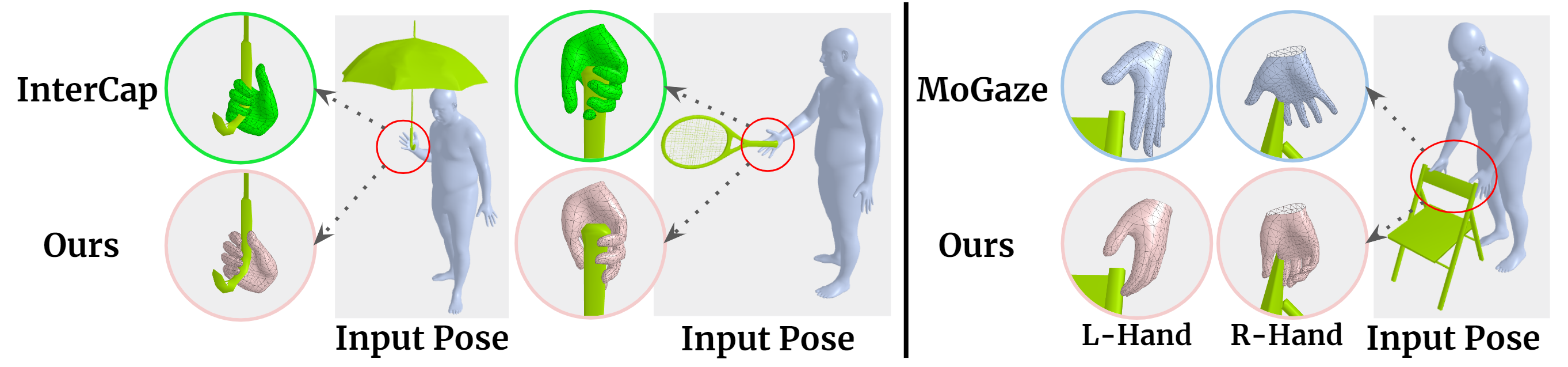}
    \caption[Generated grasps from \modelname on unseen large objects]{\modelname's performance to generate hand grasps for large objects. 
            We generate hand poses on the unseen large objects from Intercap (left) and MoGaze (right) datasets. These objects have larger 3D structures compared to the 3D objects during training, however, our hand sensors are not distracted by the extended objects due to their locality. Thus, \modelname is able to generate plausible grasps for such objects.
    }
    \label{fig:large_objects}
\end{figure*}

%% file: supmat/04_qualitative_results.tex
\section{Qualitative Results} \label{sec:qualitative_sup}

In \cref{fig:results_qualitative_sup0} we show more qualitative results generated on unseen objects, using \grip. The top row shows input body and object motion, and the bottom row shows generated hand poses. We show close-ups of the generated hand poses, in single and bimanual scenarios, to show the accuracy of the generated grasps. In \cref{fig:results_qualitative_sup1} we provide results for successive frames of a motion sequence to show the consistency of the generated hand poses over time. Additionally, the results show that our method is able to refine the noisy arm poses from the InterCap dataset. 

\input{supmat/figs/results_static}
\input{supmat/figs/results_dynamic}

In \cref{fig:manipnet_scores}, we show representative scores for the ManipNet grasps from our user study. These results confirm several limitations of ManipNet which GRIP addresses these, making it easy to apply in real-world scenarios.

\input{supmat/figs/manipnet_results}

%% file: supmat/figs/results_static.tex
\begin{figure*}[hbt!]
	\centering			%
	\includegraphics[trim=000mm 000mm 000mm 000mm, clip=true,width=1.0\linewidth]{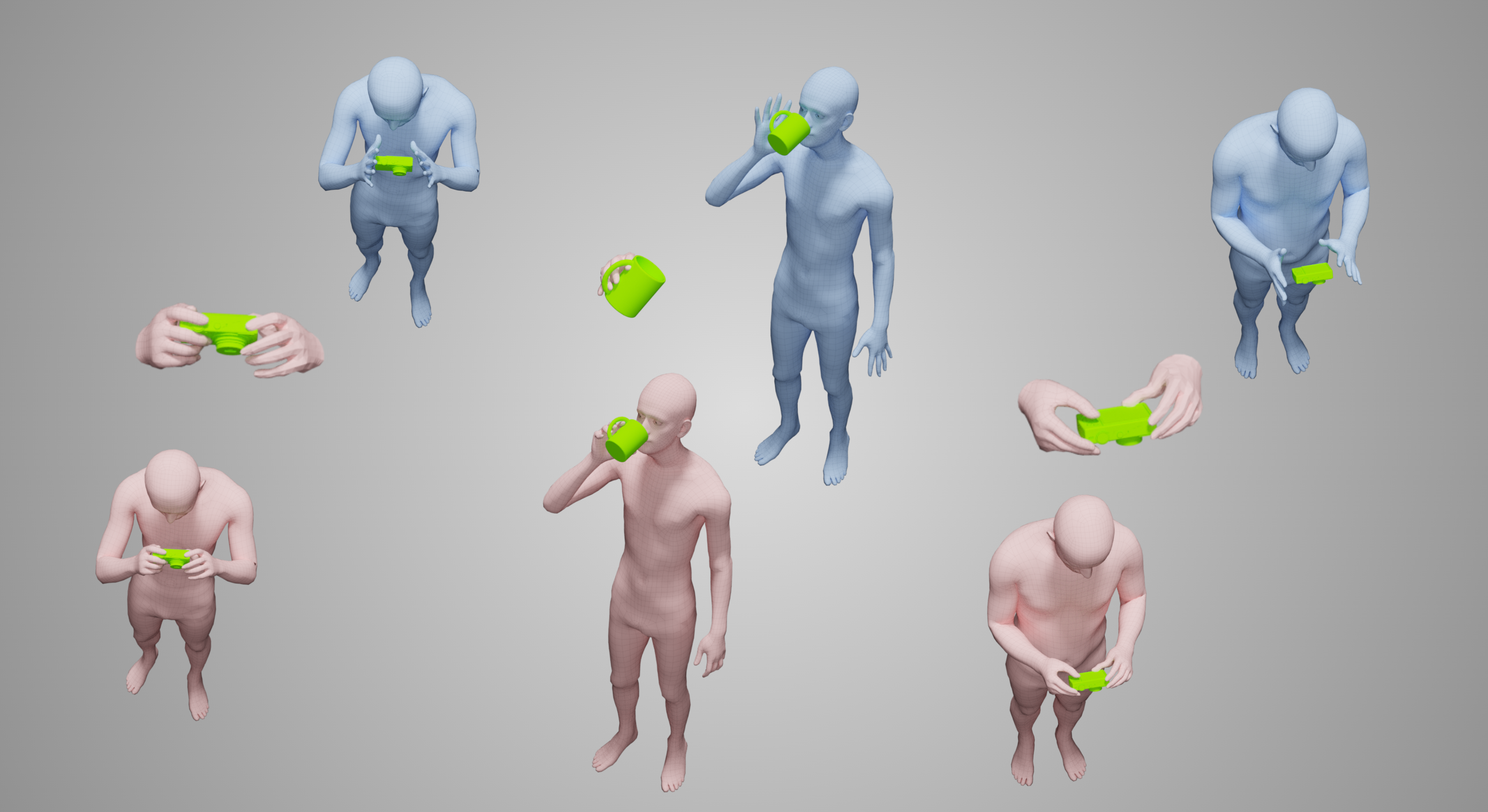}
    \caption{Generated results with \modelname for unseen objects. (Top row) input body and object, (bottom row) generated hand poses. We show close-ups of the generated hand poses in single and bimanual scenarios, to show the accuracy of the generated grasps.
    }
	\label{fig:results_qualitative_sup0}
\end{figure*}

%% file: supmat/figs/results_dynamic.tex
\begin{figure*}[hbt!]
	\centering			%
	\includegraphics[trim=000mm 000mm 000mm 000mm, clip=true,width=1.0\linewidth]{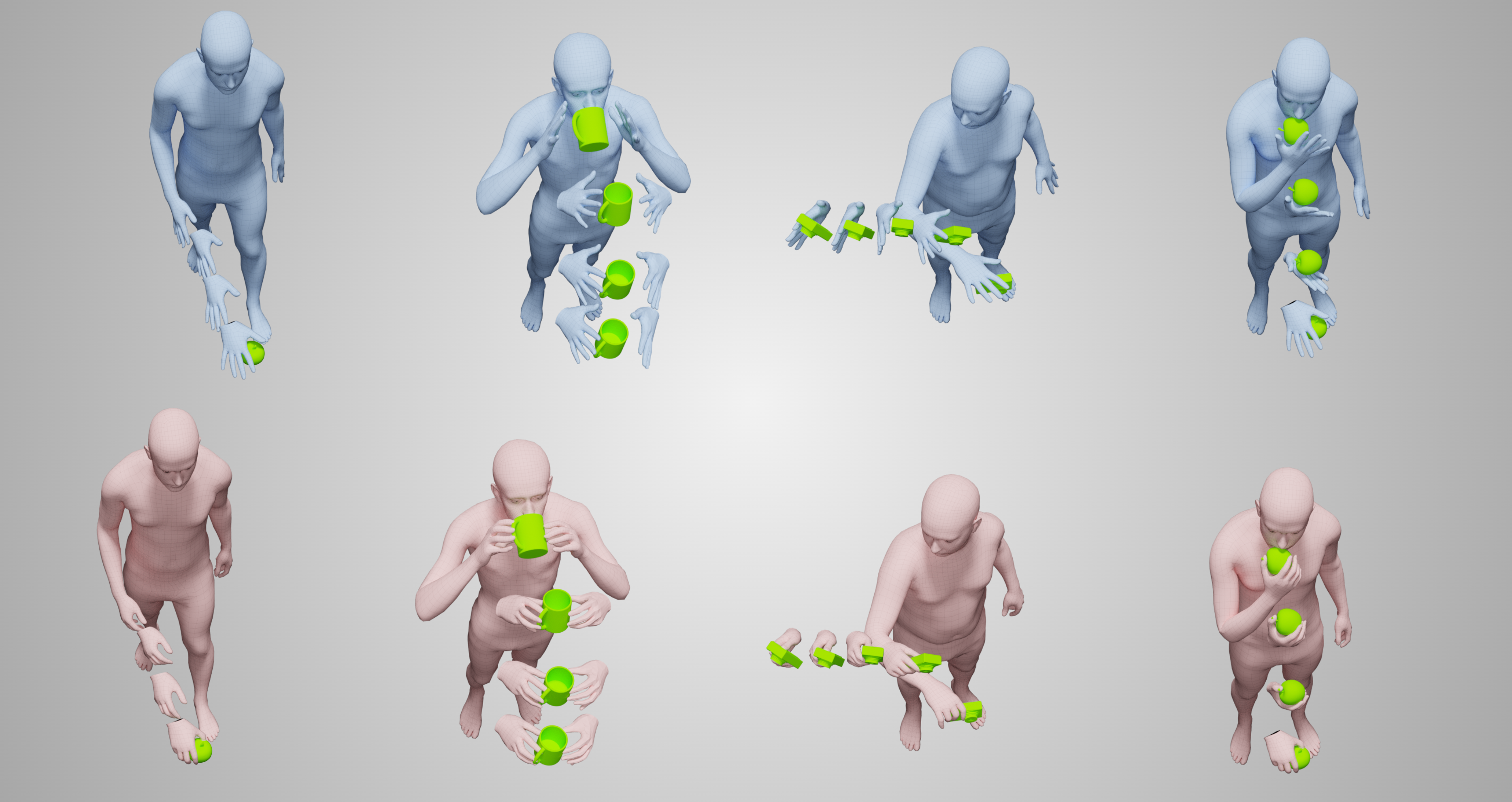}
    \caption{Generated hand motions using \modelname. (Top row) input body and object motion. (Bottom row) generated hand poses. We provide results for successive frames of the same motion to show the consistency of the generated motions over time. 
    }
	\label{fig:results_qualitative_sup1}
\end{figure*}

%% file: supmat/figs/manipnet_results.tex
\begin{figure*}[hbt!]
	\centering			%
	\includegraphics[trim=000mm 000mm 000mm 000mm, clip=true,width=1.0\linewidth]{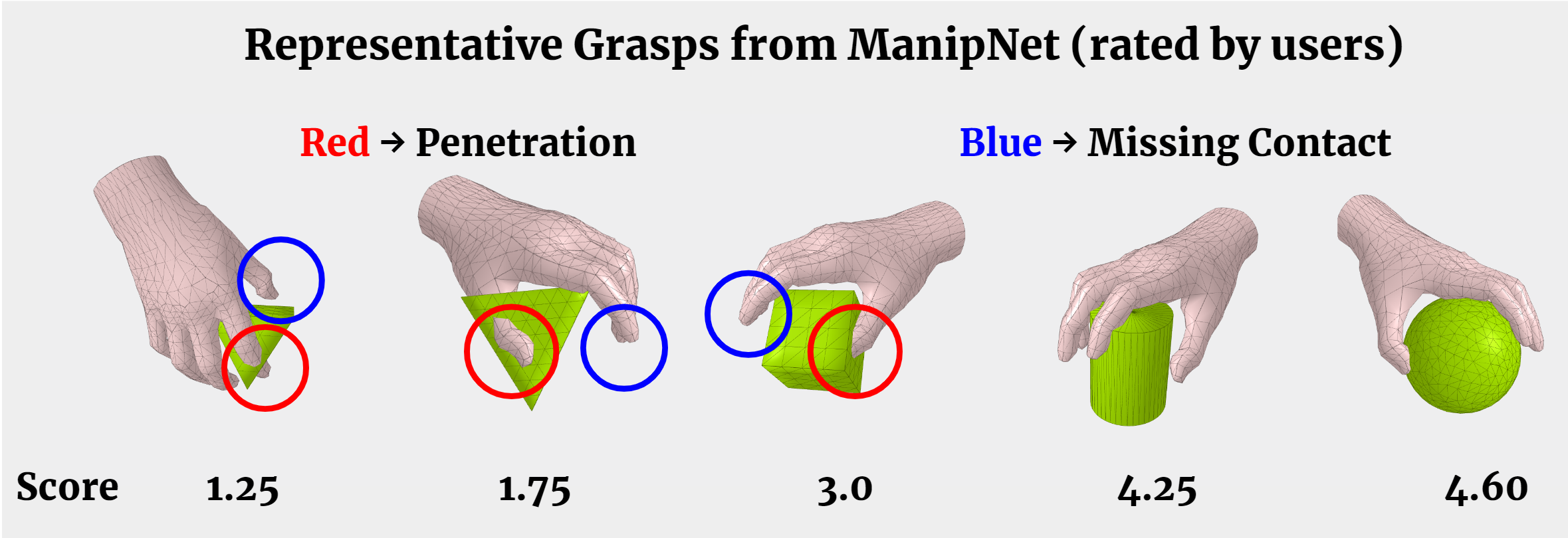}
    \caption{representative scores for ManipNet \cite{zhang2021manipnet} grasps from our user study.
    }
	\label{fig:manipnet_scores}
\end{figure*}

%% file: supmat/03_contact_heatmap.tex
\section{Grasp Analysis} \label{sec:heatmaps}

To further evaluate the quality of the generated grasps from \modelname, we compare the aggregated contact heatmaps from our method with \grab \cite{GRAB:2020}. For each motion frame in the test set, we compute the contact vertices on both hands based on their distance to the object surface, similar to \grab. We then aggregate the contact maps across all frames to compute the overall contact heatmap. Figure \ref{fig:contact_heatmap} (top) shows the contact heatmap from \grab and (bottom) shows the heatmaps for \modelname. Areas with a high likelihood of contact are shown with ``hot" (red) colors and with a low likelihood of contact are shown with ``cool'' (blue) colors. We see that \modelname contact maps follow a similar pattern to \grab, and have higher contact likelihood on the fingertips.
The similarity suggests that generated grasps exhibit similar contacts as real grasps.

\input{supmat/figs/contact_heatmap}

%% file: supmat/figs/contact_heatmap.tex
\begin{figure*}[hbt!]
	\centering			%
	\includegraphics[trim=000mm 000mm 000mm 000mm, clip=true,width=1.0\linewidth]{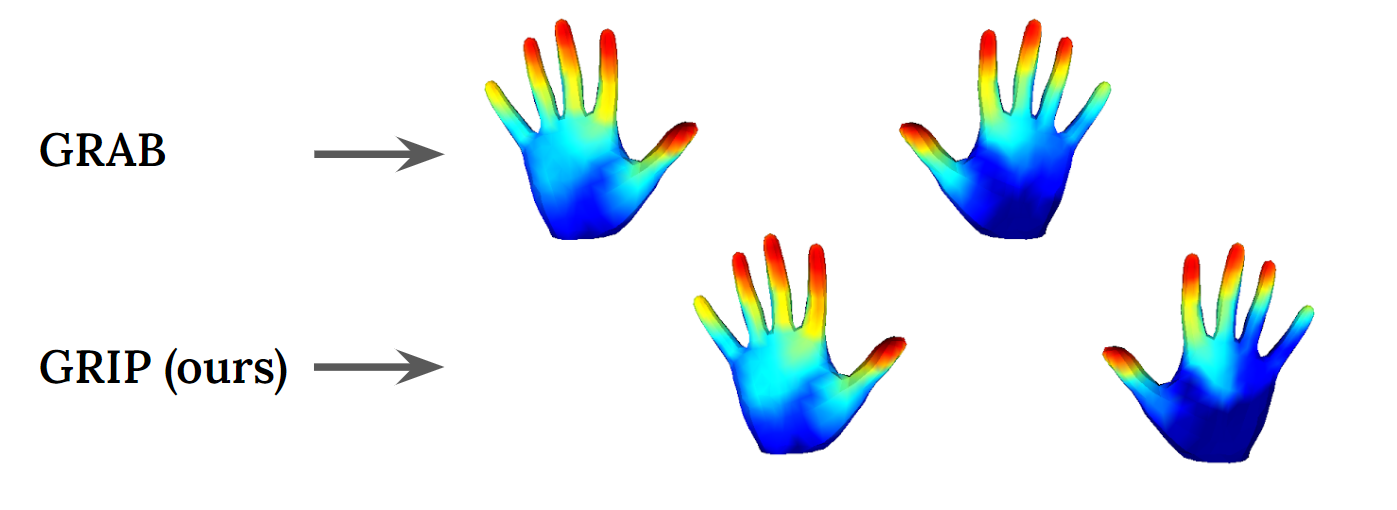}
    \caption{Comparison of the contact heatmaps from \grab and \modelname. We compute contact vertices on both left and right hand and aggregate them across all frames. Results show that \modelname contact maps are similar to \grab, which is indicative of the realism of the generated hand grasps.
    }
	\label{fig:contact_heatmap}
\end{figure*}